\documentclass[sigconf]{acmart}
\renewcommand\footnotetextcopyrightpermission[1]{} % removes footnote with conference information in first column
\usepackage{graphicx}
\usepackage{amsmath,amsfonts}
\usepackage{algorithmic}
\usepackage{graphicx}
\usepackage{textcomp}
\usepackage{hyperref}

\usepackage{multirow}
\usepackage{subfig}
\usepackage{lipsum}

\usepackage{amssymb}
\newcommand{\etal}{\textit{et al}.}
\newcommand{\ie}{\textit{i}.\textit{e}.}
\newcommand{\eg}{\textit{e}.\textit{g}.}
\newcommand{\etc}{\textit{etc}}

\newcommand\blfootnote[1]{% 
\begingroup 
\renewcommand\thefootnote{}\footnote{#1}% 
\addtocounter{footnote}{-1}% 
\endgroup 
}
\AtBeginDocument{%
  \providecommand\BibTeX{{%
    \normalfont B\kern-0.5em{\scshape i\kern-0.25em b}\kern-0.8em\TeX}}}

\setcopyright{acmcopyright}
\copyrightyear{2021}
\acmYear{2021}
\acmConference[MM '21] {Proceedings of the 29th ACM Int'l Conference on Multimedia}{October 20--24, 2021}{Virtual Event, China.}
\acmBooktitle{Proceedings of the 29th ACM Int'l Conference on Multimedia (MM '21), Oct. 20--24, 2021, Virtual Event, China}
\acmPrice{15.00}
\acmISBN{978-1-4503-8651-7/21/10}
\acmDOI{10.1145/XXXXXX.XXXXXX}

\acmSubmissionID{1223}

\begin{document}
% \begin{sloppypar}
\fancyhead{} 

%\title{One Stone Two Birds: Depth Map Super-Resolution Network Guided by Monocular Depth Estimation}
%\title{One Stone Two Birds: A Joint Learning Network of Depth Map Super-Resolution and Monocular Depth Estimation}
\title{BridgeNet: A Joint Learning Network of Depth Map Super-Resolution and Monocular Depth Estimation}


\author{Qi Tang$^{1,2}$, Runmin Cong$^{1,2,4*}$, Ronghui Sheng$^{1,2}$, Lingzhi He$^{1,2}$, Dan Zhang$^3$, Yao Zhao$^{1,2}$,\\ Sam Kwong$^4$}
\affiliation{
 \institution{$^1$Institute of Information Science, Beijing Jiaotong University, Beijing, China\\
$^2$Beijing Key Laboratory of Advanced Information Science and Network Technology, Beijing, China\\
$^3$UISEE Technology (Beijing) Co., Ltd. Beijing, China\\
$^4$City University of Hong Kong, Hong Kong, China}
}
\email{{qitang,rmcong,21120297,lingzhihe,yzhao}@bjtu.edu.cn, dan.zhang@uisee.com, cssamk@cityu.edu.hk}

% \author{Qi Tang}
% \affiliation{%
%   \institution{Beijing Jiaotong University}
%   \city{Beijing}
% %\email{qitang@bjtu.edu.cn}

% \author{Runmin Cong}
% \authornote{Corresponding author: rmcong@bjtu.edu.cn}
% \affiliation{%
%   \institution{Beijing Jiaotong University}
%   \city{Beijing}
% %\email{rmcong@bjtu.edu.cn}

% \author{Ronghui Sheng}
% \affiliation{%
%   \institution{Beijing Jiaotong University}
%   \city{Beijing}
% %\email{srh2220171815@dlmu.edu.cn}

% \author{Lingzhi He}
% \affiliation{%
%   \institution{Beijing Jiaotong University}
%   \city{Beijing}
% %\email{19112002@bjtu.edu.cn}

% \author{Dan Zhang}
% \affiliation{%
%   \institution{UISEE Technology (Beijing) Co., Ltd.}
%   \city{Beijing}
% %\email{dan.zhang@uisee.com}

% \author{Yao Zhao}
% \affiliation{%
%   \institution{Beijing Jiaotong University}
%   \city{Beijing}
% %\email{yzhao@bjtu.edu.cn}

% \author{Sam Kwong}
% \affiliation{%
%   \institution{City University of Hong Kong}
%   \city{Hong Kong}
% %\email{cssamk@cityu.edu.hk}


\begin{abstract}
%\vspace{0.3cm}
 Depth map super-resolution is a task with high practical application requirements in the industry. Existing color-guided depth map super-resolution methods usually necessitate an extra branch to extract high-frequency detail information from RGB image to guide the low-resolution depth map reconstruction. However, because there are still some differences between the two modalities, direct information transmission in the feature dimension or edge map dimension cannot achieve satisfactory result, and may even trigger texture copying in areas where the structures of the RGB-D pair are inconsistent. Inspired by the multi-task learning, we propose a joint learning network of depth map super-resolution (DSR) and monocular depth estimation (MDE) without introducing additional supervision labels. For the interaction of two subnetworks, we adopt a differentiated guidance strategy and design two bridges correspondingly. One is the high-frequency attention bridge (HABdg) designed for the feature encoding process, which learns the high-frequency information of the MDE task to guide the DSR task. The other is the content guidance bridge (CGBdg) designed for the depth map reconstruction process, which provides the content guidance learned from DSR task for MDE task. The entire network architecture is highly portable and can provide a paradigm for associating the DSR and MDE tasks. Extensive experiments on benchmark datasets demonstrate that our method achieves competitive performance. Our code and models are available at https://rmcong.github.io/proj\_BridgeNet.html.

%\vspace{0.3cm}
\end{abstract}


%\begin{CCSXML}
%<ccs2012>
%  <concept_id>10010520.10010553.10010562</concept_id>
%  <concept_desc>Computer systems organization~Embedded systems</concept_desc>
%  <concept_significance>500</concept_significance>
%  <concept_id>10010520.10010575.10010755</concept_id>
%  <concept_desc>Computer systems organization~Redundancy</concept_desc>
%  <concept_significance>300</concept_significance>
%  <concept_id>10010520.10010553.10010554</concept_id>
%  <concept_desc>Computer systems organization~Robotics</concept_desc>
%  <concept_significance>100</concept_significance>
%  <concept_id>10003033.10003083.10003095</concept_id>
%  <concept_desc>Networks~Network reliability</concept_desc>
%  <concept_significance>100</concept_significance>
%</ccs2012>
%\end{CCSXML}
%
%\ccsdesc[500]{Computer systems organization~Embedded systems}
%\ccsdesc[500]{Computing methodologies~Computer vision problems}






\maketitle
\blfootnote{*Corresponding author.}
\vspace{-0.5cm}
\section{Introduction}
\label{sec:introduction}

When understanding a scene, people can not only perceive its appearance (\eg, color, texture), but also capture the depth information to generate the stereo perception. Better scene understanding can facilitate many research areas such as autonomous navigation \cite{DBLP:conf/iros/KerlSC13}, 3D reconstruction \cite{DBLP:journals/pami/ImHCJJK19}, \etc, all of which rely on high-quality depth information. The emergence and popularization of portable consumer-grade depth cameras, such as Microsoft Kinect and Lidar, provide great convenience for quickly acquiring the depth of scene \cite{crm2019tcsvt}. However, due to the limitation of the imaging capabilities of depth cameras, the resolution of depth map is usually limited, let alone paired with high-resolution color image. Facing with the urgent demand for high-quality depth map in applications \cite{DBLP:journals/pami/GeLYT19,SilbermanHKF12,DBLP:journals/corr/abs-1907-06781,crm2020tc,DPANet,crm2019tmm}, depth map super-resolution (SR) technology as a solution has attracted more and more attention.

\begin{figure}[!t]
\centering
\centerline{\includegraphics[width=0.46\textwidth]{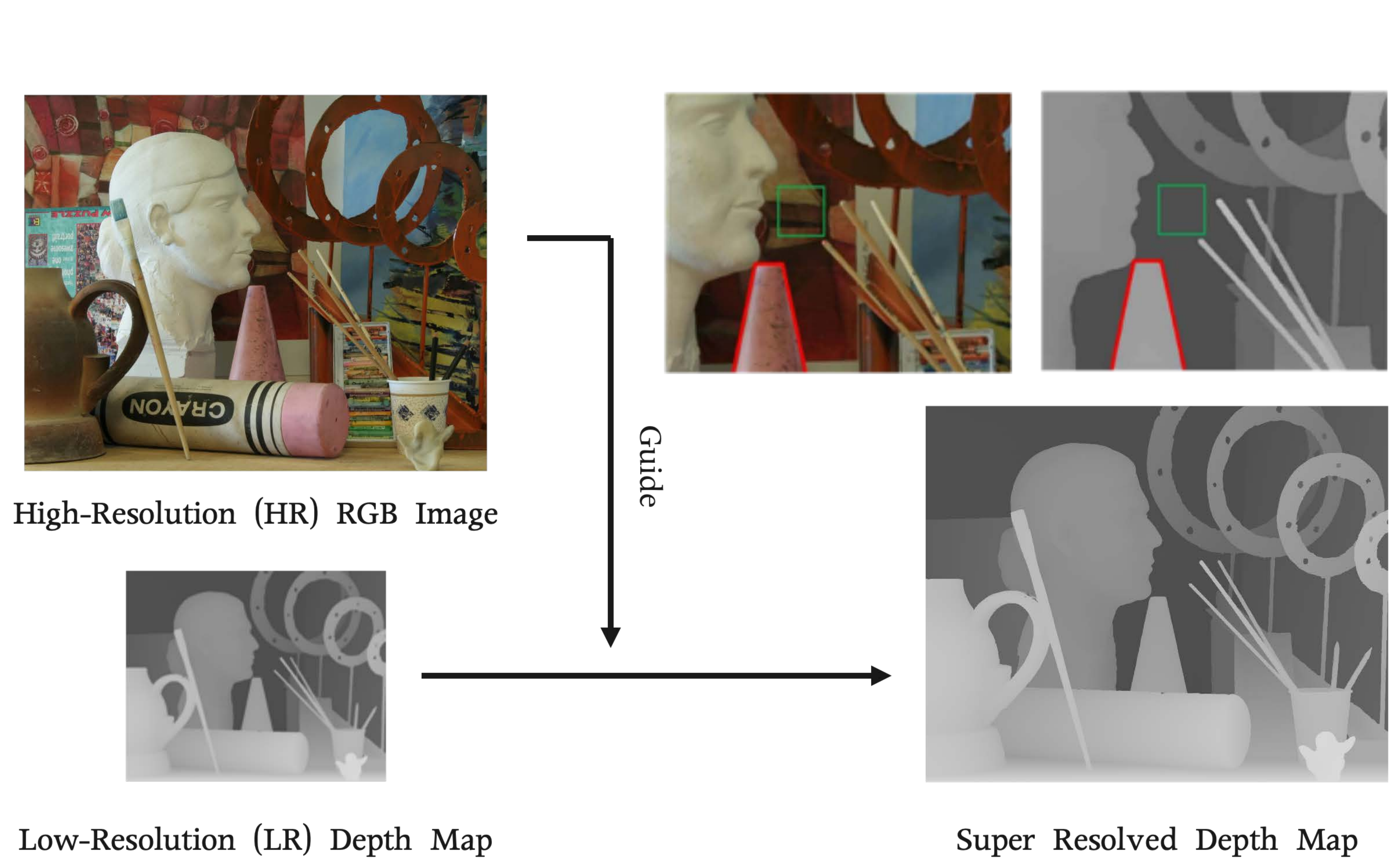}}
\caption{Color-guided depth map super-resolution. In the red trapezoidal region, the two have the same edge structure. While in the green rectangular region, there are more texture changes in the color image, but no corresponding texture structure in the depth map.}
\label{fig:im_sample}
\vspace{-0.5cm}
\end{figure}

Depth SR aims to super-resolve a low-resolution (LR) depth map to a high-resolution (HR) depth map. It not only needs to generate the high-frequency counterpart (the high-frequency information of the image is the region where the gray level changes rapidly, such as the object edge) of the depth map from the degraded LR depth map, but also needs to effectively suppress the random noise and blur phenomenon in the imaging process. It is difficult to recover accurate HR depth map using artificially constructed filters or objective functions in the traditional filter-based methods \cite{LuSMLD12,0001TT13} and optimization-based methods \cite{ParkKTBK11,DBLP:conf/iccv/FerstlRRRB13}. In recent years, with the success and application of deep learning technology, many works have also verified its role in depth SR task \cite{HuiLT16,WenSLLF19,DBLP:journals/corr/abs-2104-06174}, which can reconstruct HR depth map by automatically learning stronger representations from data. In practical application scenarios, high-resolution color image is easily obtained and has strong structural similarity with the depth map \cite{crm2019tc,crm2018tip,crm2016spl}, thus it can provide some guidance information for depth SR. We call this type of method color-guided depth super-resolution. Existing color-guided depth SR methods usually require an extra branch to acquire rich guidance information \cite{LutioDWS19}, and then use it to guide the hierarchical feature learning of the depth branch. However, the edges of the color image do not always coincide with the depth map. For example, as shown in Figure \ref{fig:im_sample}, the green rectangular region has a complex texture structure on the color image, but this area is displayed as a smooth region in the corresponding depth map. In this way, if we simply pass the RGB features or the extracted RGB edge features to the depth branch, it is easy to trigger issues such as texture copying and depth bleeding \cite{lcy2020tc,eccv20}. Therefore, how to effectively explore the high-resolution color information is very important to depth SR task.

Before searching for a solution to the above problem, let us turn our attention to another depth-related task, namely monocular depth estimation, aiming to map a scene from the photometric representation to the geometrical representation. In layman's terms, the input of the monocular depth estimation is an RGB image, and the output is the estimated depth map. The two tasks are naturally related. The first is that the training datasets for these two tasks can be shared. If the two models are put in a unified framework for training, there is no need to introduce additional supervision labels (\eg, semantic labels). In addition, orientated by the task of monocular depth estimation, the features learned from RGB image are more suitable for guiding the depth SR, because it can realize this kind of cross-modality information transformation in the continuous training and learning process. In summary, the joint learning of depth map super-resolution reconstruction and monocular depth estimation can achieve better color guidance without increasing the supervision information.

Motivated by the above analyses, we propose a joint learning network of depth super-resolution (DSR) and monocular depth estimation (MDE), namely  BridgeNet, focusing on achieving better depth SR by effectively bridging the two tasks. To this end, we design two subnetworks based on the encoder-decoder architecture, \ie, DSRNet and MDENet, which work together in a multi-task learning manner. Moreover, two different bridges in the encoder stage and decoder stage are designed to achieve differentiated guidance of two subnetworks. The encoder of MDENet learns the multi-level features oriented towards depth map from RGB image, which is suitable for guiding the depth SR branch. Thus, we propose a high-frequency attention bridge (HABdg) to learn the high-frequency information of MDENet to guide the depth SR branch. The feature decoders of MDENet and DSRNet are used to further extract task-oriented features for depth estimation and depth super-resolution. In contrast, the MDE task is more difficult than the DSR task because of its scale ambiguity. Therefore, following the principle of simple task guiding difficult task, we propose a content guidance bridge (CGBdg) to let DSRNet provide content guidance for MDENet in the depth feature space. In addition to associating the two tasks at the model design level, we also constrain them in terms of loss function, in the hope that the two subnetworks can promote each other.

To sum up, the contributions of this work are as follows:
\begin{itemize}
  \item This work attempts to associate the depth map super-resolution and monocular depth estimation in a joint learning framework to boost the performance of depth map super-resolution, including a depth super-resolution network (DSRNet), a monocular depth estimation network (MDENet), and two joint learning bridges. Our entire network architecture is highly portable and can provide a paradigm for associating the DSR and MDE tasks. Moreover, different from other multi-task learning, no additional labels are required.
  \item The high-frequency attention bridge (HABdg) in the feature encoding stage transfers the RGB high-frequency information learned from MDENet to DSRNet, which can provide color guidance information closer to the depth modality. Following the principle of simple task guiding difficult task, we switch the guiding roles of the two tasks in the feature decoding stage, and propose the content guidance bridge (CGBdg) to provide the content guidance learned from DSRNet for MDENet.
  \item Without the introduction of additional supervision labels, our method achieves competitive performance on benchmark datasets.

\end{itemize}
 \vspace{-0.15cm}
 
\begin{figure*}[!htbp]
\centering
\centerline{\includegraphics[width=1\textwidth]{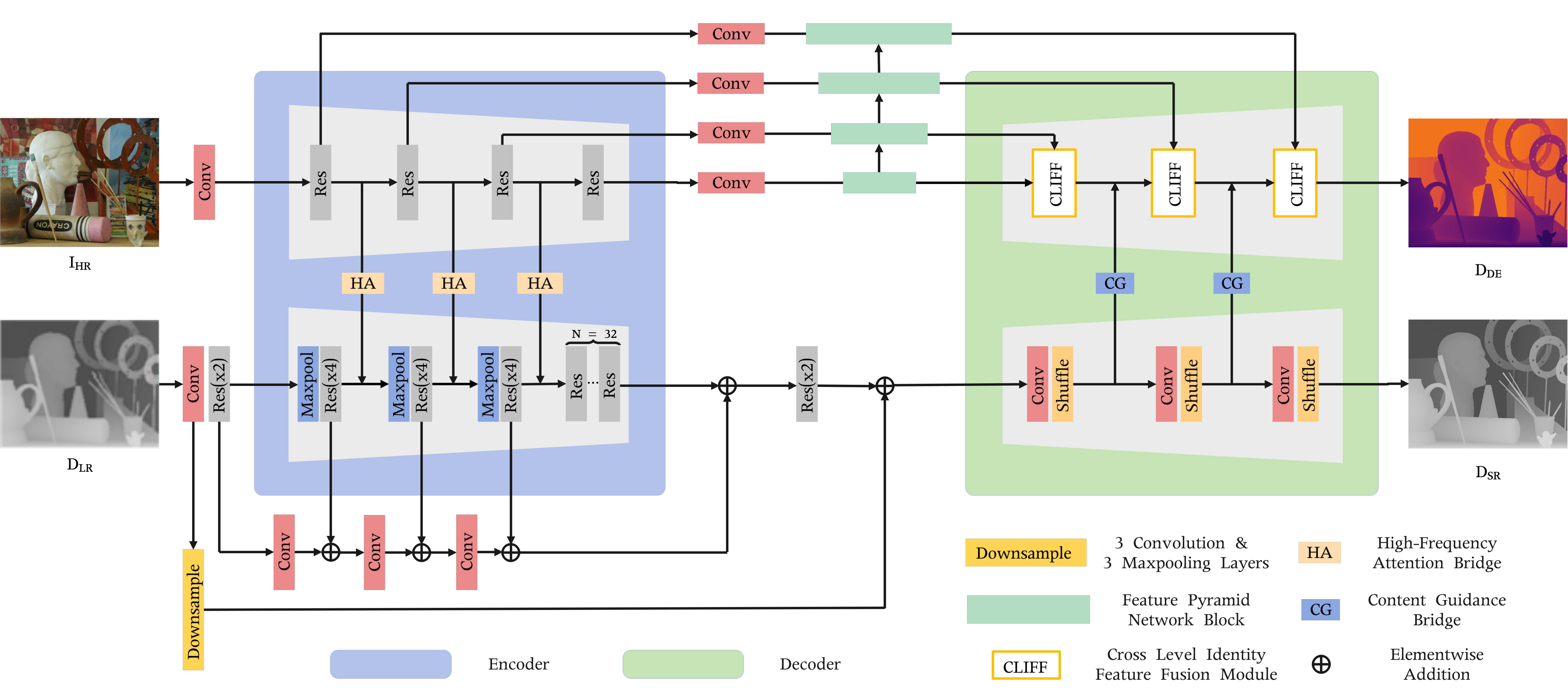}}
\caption{Architecture of the proposed BridgeNet, which consists of a depth super-resolution subnetwork (DSRNet), a monocular depth estimation subnetwork (MDENet), a high-frequency attention bridge (HABdg), and a content guidance bridge (CGBdg). The encoder-decoder structure at the top is the MDENet, and the bottom encoder-decoder structure corresponds to the DSRNet. The HABdg works between the feature encoders of the two subnetworks, focusing on passing the high-frequency color guidance obtained from MDENet to DSRNet. On the contrary, CGBdg works on the decoder side and is used to provide the MDENet with content guidance information learned from the DSRNet.}
\label{fig:fig_2}
\vspace{-0.3cm}
\end{figure*}

\section{Related Work}
%\vspace{0.3cm}

\noindent
\textbf{Depth Map Super-Resolution.} Due to the structural similarity between color image and depth map, numerous methods have been proposed to use color information to guide the reconstruction of degraded LR depth map. Zhao \etal~ \cite{DBLP:journals/corr/abs-1708-09105} proposed a texture-depth conditional generation confrontation network to learn the structural similarity between texture images and low-resolution depth maps. Hui \etal~\cite{HuiLT16} designed a multi-scale guided convolutional network for depth SR, which uses the rich hierarchical texture feature to eliminate the blurring phenomenon after depth map reconstruction. Zuo \etal~\cite{DBLP:journals/isci/ZuoFYSW19} proposed a texture-guided enhanced residual dense network and a multi-scale fusion residual network to explore how to use the multi-scale guidance information provided by texture images to gradually guide the up-sampling and recovery of depth map. Ye \etal~\cite{DBLP:journals/tip/YeSWYXLL20} proposed a progressive multi-branch aggregation network by using the multi-branch fusion method to gradually restore the degraded depth map. Guo \etal~ \cite{DBLP:journals/tip/GuoLGCFH19} used the U-Net structure to encode the interpolated depth image, and fused the texture features of the corresponding scale during the decoding process.%

\noindent
\textbf{Monocular Depth Estimation.} Monocular depth estimation is a typical inverse problem, since we are attempting to recover some unknowns given insufficient information to fully specify the solution. Compared with the depth estimation task based on the left and right views, the monocular depth estimation has broader practical application prospects, but the current estimation performance is still very limited. Eigen \etal~ \cite{DBLP:conf/nips/EigenPF14} estimated depth from a monocular image by using a convolutional neural network with two scales, which establish a precedent for deep learning based MDE method. To achieve up-sampling, Laina \etal ~\cite{DBLP:conf/3dim/LainaRBTN16} used a deeper residual network and small convolutions rather than large convolutions. Cao \etal ~\cite{DBLP:journals/tcsv/CaoWS18} improved performance by discretizing the original continuous depth into a fixed number of ranges of predetermined width and transforming the depth regression task into a classification task.

\noindent
\textbf{Depth-Oriented Multi-Task Learning.} The purpose of multi-task learning is to boost the performance of a specific task by combining other related tasks. The simplest and most common way is to combine the two tasks through the loss function. But this is often not optimal because of the lack of interaction in network design. At present, many tasks related to depth processing have adopted the multi-task learning strategy.
Zhang \etal~\cite{DBLP:conf/eccv/ZhangCXJLY18} proposed Task-Recursive Learning (TRL) framework recursively refine the results of semantic segmentation and monocular depth estimation tasks based on serializing the problems as a task-alternate time sequence.
%Those methods only learn the two tasks without specifically addressing the constraints that exist between them.
He \etal~\cite{DBLP:journals/corr/abs-2101-07422} proposed a SOSD-Net for simultaneous monocular depth estimation and semantic segmentation based on the geometric relationship of these two tasks.
However, the correlation between the tasks modeled by these methods is still weak, and additional labels (such as semantic labels) are required for training.
Sun \etal~\cite{Sun2021cvpr} proposed a knowledge distillation method to enable MDE to help DSR better understand the structure of the scene in the training process, thereby demonstrating effectiveness of MDE in improving DSR performance. By comparing the average pixel error of the two tasks in the training process to determine the interaction method of the two tasks, this does not directly explore the correlation between DSR and MDE.

In this paper, we propose a joint learning network of MDE and DSR to achieve better performance of DSR. Different form the previous work \cite{Sun2021cvpr}, when designing the interaction between the two subnetworks, we adopt a more explicit guidance mode. In the feature encoder, we let the MDE task provide high-frequency guidance information for the DSR task through the HABdg, so that the color guidance provided will be closer to the depth modality. In the feature decoder, we follow the principle of simple task guiding difficult task, and use DSR branch to provide content guidance for depth estimation branch via the CGBdg. Moreover, our method surpasses the work of \cite{Sun2021cvpr} in performance of multiple evaluation indicators.

\vspace{-0.1cm}
\section{Proposed Method}

\subsection{Network Architecture}
%\vspace{0.2cm}

Figure \ref{fig:fig_2} depicts the overall architecture of the proposed network, which consists of two subnetworks (\ie, DSRNet and MDENet) and two bridges (\ie, HABdg and CGBdg). The DSRNet and MDENet are equipped into a unified framework to achieve depth super-resolution and monocular depth estimation jointly, and the HABdg and CGBdg are respectively applied to the encoder and decoder stages to link these two tasks together.
Given a set of HR RGB-D pairs $\{I_{HR}^{(n)},D_{HR}^{(n)}\}_{n=1}^N$ and the corresponding LR depth maps $\{D_{LR}^{(n)}\}_{n=1}^N$ as training data, where $N$ is the number of training images. Moreover, LR depth maps are interpolated to the size of HR depth maps. Our model takes the LR depth map ($D_{LR}$) and the corresponding HR RGB image ($I_{HR}$) as inputs to train the DSRNet and MDENet simultaneously. The super-resolved depth map ($D_{SR}$) is the main output of our network, and we also output the estimated depth map ($D_{DE}$) as the auxiliary.

\textbf{Monocular Depth Estimation Subnetwork (MDENet).}
The encoder-decoder architecture has achieved great success in the monocular depth estimation. In our model, we follow the encoder-decoder network architecture used in \cite{DBLP:conf/eccv/WangZWLR20} as our MDENet, which is mainly composed of three parts, that is, feature extractor, feature pyramid, and depth prediction. The feature extractor learns the multi-level features from the input RGB image, and the feature pyramid module propagates the high-level features to low-level features and generates the refined multi-level features. During the feature decoding, the cross-level identity feature fusion (CLIFF) module is used to progressively fuse the refined multi-level features and achieve depth estimation, where the interpolated high-level features and low-level features are fed as the inputs. It will refine the low-level features by multiplying them with the high-level features, so that accurate responses in the low-level features are further strengthened. Finally, the high-level features, original and refined low-level features are selected through two convolutional layers, thereby obtaining the most beneficial features to the MDE.

\begin{figure*}[!t]
\centering
\centerline{\includegraphics[width=0.9\textwidth]{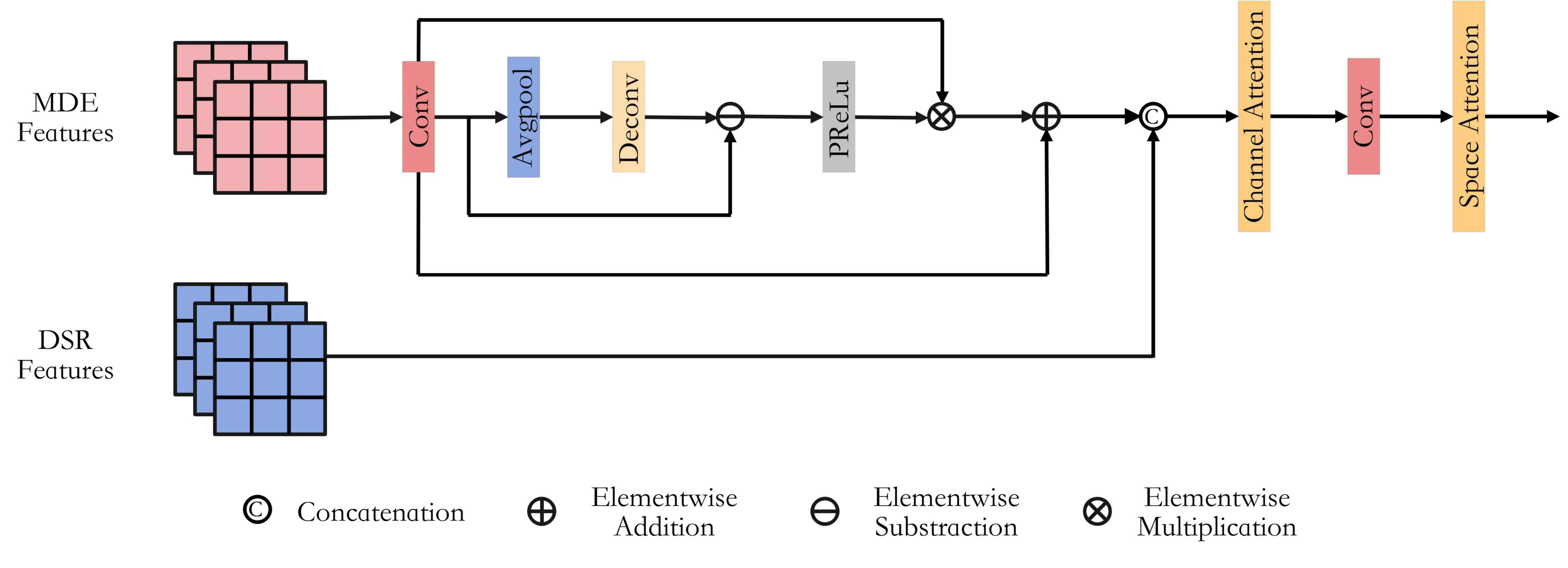}}
\caption{Illustration of HABdg. We first learn the high-frequency attention from the encoder features of the MDENet. Then, it is used to weight the original features to obtain the refined guidance features. After cascading with the features of the corresponding layer of DSRNet, the final output features (\ie, the features of feeding into the next DSRNet encoder layer) are obtained through the CA and SA mechanisms.}
\label{fig:fig_4}
\vspace{-0.3cm}
\end{figure*}

\textbf{Depth Super-Resolution Subnetwork (DSRNet).}
Following the architecture of face super-resolution network \cite{DBLP:conf/aaai/YinRZF20}, we also take the encoder-decoder network as our baseline of DSRNet. The encoder stage comprises a convolutional layer, a residual block, and three consecutive transformation modules, each of which module consists of a max-pooling layer and four residual blocks in series. Considering the positive effect of deeper network on super-resolution,  we use some stacked residual blocks to further enhance the feature representation. Then, in order to  recover the fine structure and tiny objects, we introduce the multi-scale strategy to guide the top-level features by further fusing the encoder features of the middle layers. Moreover, the shallow features in the encoder are fed into the Downsample block to generate the low-frequency features, and are combined with encoder features via a long skip connection, which forces the network to focus on learning high-frequency information in depth SR. The Downsample block consists of three convolution layers, each of which is followed by a max-pooling layer. During feature decoding, we use three identical modules connected sequentially, including a convolutional layer and a pixel shuffle layer. Feature maps of each layer are upsampled twice in resolution, and a $1 \times 1$ convolutional layer is finally applied to reconstruct the HR depth map.

\textbf{Joint Learning Strategy.}
Depth super-resolution and monocular depth estimation have a natural correlation, and they can be trained under the supervision of the same dataset. Therefore, the joint learning of these two tasks is first manifested in the joint optimization of the loss function.
Different from other multi-task learning methods whose loss function is a weighted sum of all branches, we assign different optimizers for the loss functions of DSR and MDE, respectively. The reason for this is that the learning difficulty of DSR and MDE is quite different, resulting in different convergence speeds of two tasks, and it is difficult to find a suitable weight setting to ensure that both tasks achieve the best performance. Therefore, in the design of the loss function, we separately optimize the parts related to DSR and MDE. Thus, their losses are defined as follows:
\begin{align}
	\mathcal{L}_{DSR}&=||D_{SR}-D_{HR}||_1\\[0.5em]
	\mathcal{L}_{MDE}&=||D_{DE}-D_{HR}||_1
\end{align}
where $\mathcal{L}_{DSR}$ and $\mathcal{L}_{MDE}$ are pixel-wise $L_1$ loss for the task of DSR and MDE, respectively.

In addition to the loss constraint, we also elaborately design two bridges to associate the two tasks in the encoder and decoder stages respectively to achieve mutual benefit and common progress.
One is the high-frequency attention bridge (HABdg) in the encoder, which uses the excellent color feature learning ability of the MDENet to provide guidance for the depth map super-resolution.
The other is the content guidance bridge (CGBdg) in the decoder. Considering the difficulty of the two prediction tasks, the decoding features of DSRNet can provide effective content guidance for depth estimation, thereby improving the effect of depth estimation. In the following sections, we will stress the principles and details of these two bridges.

\subsection{High-Frequency Attention Bridge}
%\vspace{0.2cm}

First, let us consider the correlation of the encoder parts of the two subnetworks. Recalling the previous color-guided depth SR methods, the guidance methods of color image mainly include direct corresponding feature guidance \cite{HuiLT16,DBLP:journals/tip/GuoLGCFH19,LutioDWS19} or edge detail guidance \cite{DBLP:journals/tmm/WangXCZSY20, DBLP:conf/icassp/YeDL18}. Although the color image and depth map have strong structural similarities, the abundant texture and edge of the color image are not always consistent with the depth map, so that the direct feature guidance or edge guidance may result in texture copying and depth bleeding. Looking back at the monocular depth estimation task, its purpose is to start from an RGB image, mapping photometric representation to geometrical representation, and then generate the depth map. Therefore, the features of color image provided by the depth estimation encoder are closer to the feature representation of depth modality, which can avoid the noticeable artifacts when providing high-frequency information guidance for DSR. This is why we use the depth estimation branch to guide the depth super-resolution branch in the encoder stage.

After determining the guiding direction of information transmission, the next question is how to effectively implement it. The simplest and most intuitive way is to pass the corresponding-layer features of the MDENet directly to the DSRNet through concatenation or addition. But this is obviously not a wise way.
In the MDENet encoder, as the network deepens, the feature resolution gradually decreases, among which high-level features have rich semantic information, while low-level features have more structural information. Since the LR depth map contains less high-frequency information, it is more important that the HR color image can provide high-frequency information (such as the edge details) rather than the semantic information of the image. Motivated by this, we design a high-frequency attention bridge, which is specially used to learn the high-frequency information from the MDENet to guide the depth SR branch. The pipeline of the HABdg is shown in Figure \ref{fig:fig_4}.

Specifically, we first use the average-pooling and deconvolution operations to blur the original features of the MDENet, which is formulated as:
\begin{align}
	F_{blurred}^i=deconv(avgpool(F_{MDE}^i)),
\end{align}
where $F_{blurred}^i$ is the obtained blurred features of the $i^{th}$ layer, $F_{MDE}^i$ denotes the encoder features of the $i^{th}$ layer in MDENet, $avgpool(\cdot)$ and $deconv(\cdot)$ are the average-pooling and deconvolution operations, respectively.

Then, we calculate the the high-frequency information by subtracting between the original features and the blurred ones, thereby generating the high-frequency attention weight:
\begin{align}
	A_{hf}^i=PRelu(F_{MDE}^i-F_{blurred}^i),
\end{align}
where $A_{hf}^i$ denotes the high-frequency attention of the $i^{th}$ layer, and $PRelu(\cdot)$ represents the parametric rectified linear unit. Next, we use the high-frequency attention to refine the original features $F_{MDE}^i$ through residual connection and obtain the refined guidance features:
\begin{align}
	F_{hg}^i=F_{MDE}^i+ A_{hf}^i  \cdot F_{MDE}^i,
\end{align}
where $F_{hg}^i$ represents the refined guidance features of the $i^{th}$ layer.
The intuition is to highlight the high-frequency in the original features, so that the LR depth map can better yield HR counterpart during feature fusion.

With the refined encoder guidance features of MDENet, we first concatenate them with the corresponding encoder features of DSRNet to generate the composition features $F_{comp}^i$. Such a simple feature fusion will have a lot of redundancy in the spatial dimension and channel dimension, thus we introduce an attention block including a channel attention \cite{DBLP:conf/eccv/WooPLK18} and a spatial attention  \cite{DBLP:journals/tip/SindagiP20} to enhance the fused features. The channel attention learns the importance of each feature channel, and spatial attention highlights the important spatial locations in the feature map. These processes can be formulated as:
\begin{align}
	F_{comp}^i=[F_{DSR}^i, F_{hg}^i],
\end{align}
\begin{align}
    F_{ha}^{i}=SA(conv_{1\times 1}(CA(F_{comp}^i))),
\end{align}
where $F_{DSR}^i$ denotes the encoder features of the $i^{th}$ layer in DSRNet, $CA$ and $SA$ are channel attention and spatial attention blocks, respectively, $conv_{1 \times 1}$ is the convolutional layer with the kernel size of $1 \times 1$, and $[\cdot ~, \cdot]$ denotes the channel-wise concatenation operation. The output features $F_{ha}^{i}$ will be used as the input of the next layer in DSRNet.

\subsection{Content Guidance Bridge}
%\vspace{0.2mm}

\begin{figure}[!t]
\centering
\centerline{\includegraphics[width=0.44\textwidth]{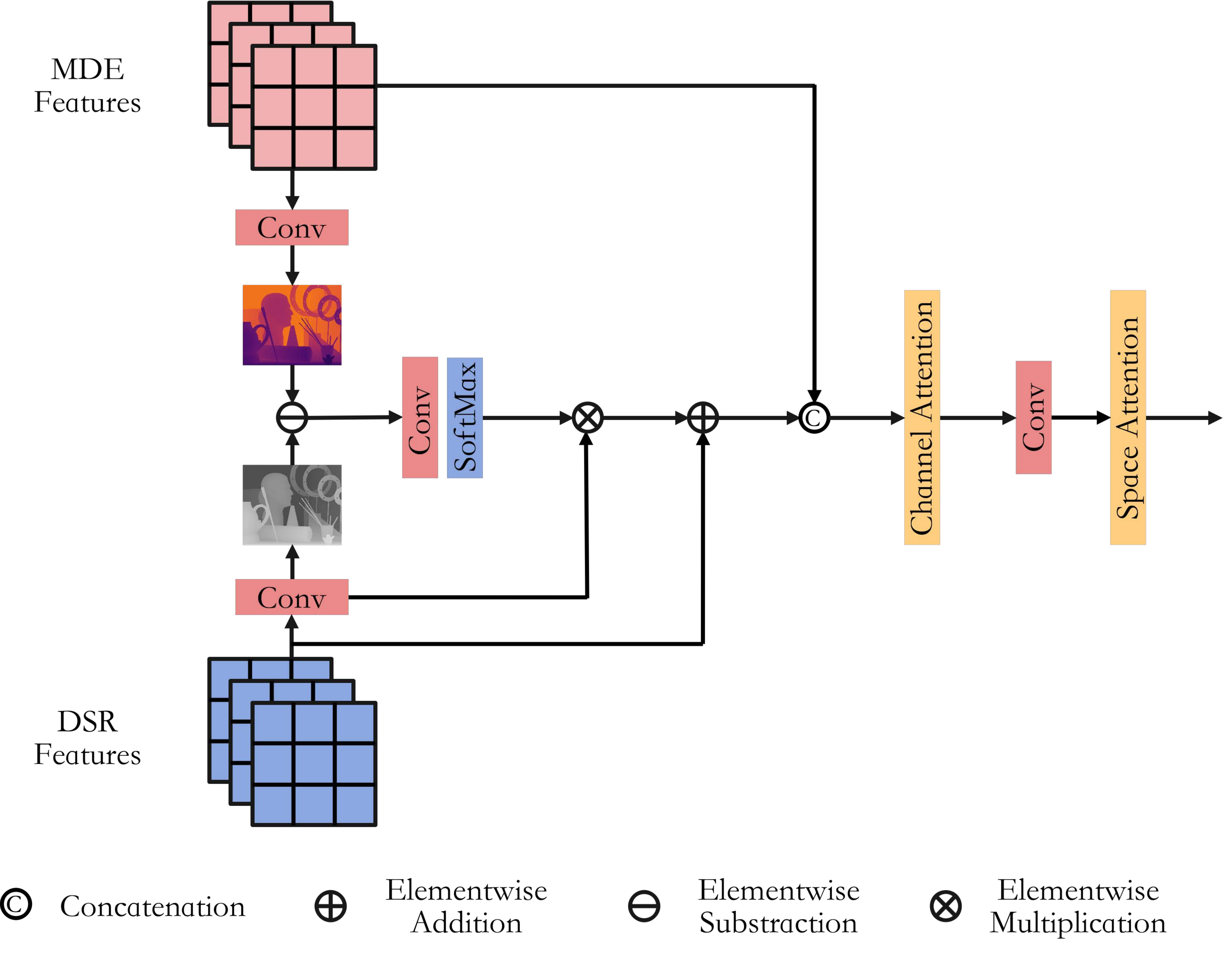}}
\caption{Illustration of CGBdg. We first calculate the difference map between estimated depth map and super-resolved depth map, and then learn the difference weight through a convolution operation and softmax activation. Applying the difference weight to the depth SR encoder features to generate the content guidance for the depth estimation branch. Finally, the depth estimation features and content guidance are concatenated, and fed into the channel attention and spatial attention modules to produce the output features.}
\label{fig:fig_6}
\vspace{-0.4cm}
\end{figure}

For the feature decoding stages of MDENet and DSRNet, their roles are to further extract the task-oriented features for depth estimation and depth super-resolution. In this way, we can obtain the corresponding depth maps of the two subnetworks, either estimated or super-resolved. Compared with these two tasks, monocular depth estimation is known as the ill-posed inverse problem because of the scale ambiguity \cite{DBLP:conf/nips/EigenPF14}. What it means is that many 3D scenes observed in the world can indeed correspond to the same 2D plane, that is, they are not in a one-to-one correspondence. As a result, training a model that maps well from RGB to depth is a very difficult task. Although the depth map super-resolution reconstruction is also an ill-posed problem, it learns the mapping in the same domain and focuses on restoring details, which is relatively simpler than monocular depth estimation.  Therefore, due to the large gap between the performance of the two tasks, the decoder features of MDENet is no longer appropriate to providing guidance information for the decoding stage of DSRNet. Following the principle of simple task guiding difficult task, we exchange the guiding roles of the two subnetworks in the decoding stage, that is, let DSRNet provide content guidance for MDENet in the depth feature space. The detailed structure is illustrated in Figure \ref{fig:fig_6}.

As mentioned earlier, we can obtain two depth maps according to the decoder features of the two subnetworks. Specifically, we apply a convolutional layer with the kernel size of $1\times1$ to the decoder features, obtaining a super-resolved depth map and a estimated depth map:
\begin{align}
	M_{DSR}^i=conv_{1 \times 1}(Fd_{DSR}^i)
\end{align}
\begin{align}
	M_{MDE}^i=conv_{1 \times 1}(Fd_{MDE}^i)
\end{align}
where the $Fd_{DSR}^i$ and $Fd_{MDE}^i$ are the decoder features of the $i^{th}$ layer in the DSRNet and MDENet respectively, $M_{DSR}^i$ and $M_{MDE}^i$ denote the depth maps of the $i^{th}$ layer predicted by the DSRNet and MDENet respectively.

Then, we calculate the difference map between the estimated depth map $M_{MDE}^i$ and super-resolved depth map $M_{DSR}^i$. The difference map highlights those positions in the estimated depth map that need to be further optimized relative to the super-resolved depth map. As the network is trained, we hope that this difference will become smaller and smaller. Based on this, we learn the difference weight by applying a convolution operation and softmax activation to the difference map, and further generate the content guidance for the depth estimation branch. The above procedures are formulated as:
\begin{align}
	W_{diff}^i=softmax(conv_{1 \times 1}(M_{DSR}^i-M_{MDE}^i))
\end{align}
\begin{align}
	F_{cg}^i=Fd_{DSR}^i+W_{diff}^i*Fd_{DSR}^i
\end{align}
where $W_{diff}^i$ denotes the difference weight, $F_{cg}^i$ is the content guidance of the $i^{th}$ layer, and $softmax$ represents the softmax activation. Finally, similar to the HABdg, we still use the attention block to optimize the concatenation features (\ie, $F_{con}^i=[Fd_{MDE}^i, F_{cg}^i]$), so as to obtain the features that are fed to the next layer of decoding block in MDENet.

\section{Experiments}
%\vspace{0.2cm}
\subsection{Training and Implementation Details}
%\vspace{0.2cm}
We collect 36 RGB-D pairs from Middlebury dataset (6, 21, 9 images from 2001 \cite{DBLP:journals/ijcv/BakerSLRBS11}, 2006 \cite{DBLP:conf/cvpr/HirschmullerS07}, and 2014 \cite{DBLP:conf/dagm/ScharsteinHKKNWW14} datasets, respectively) for training, and 6 RGB-D pairs of the Middlebury 2005 \cite{DBLP:conf/cvpr/ScharsteinP07} dataset for testing. Another training and testing dataset is NYU v2 dataset \cite{SilbermanHKF12}. Following the common splitting method \cite{DBLP:conf/eccv/LiHA016}, we use the first 1000 pairs as training data, and evaluate on the last 449 pairs. The RGB-D pairs from training and testing are all normalized to the range of [0, 1].

Following the previous method \cite{HuiLT16}, sufficient patches are cropped by dividing each HR image into a regular grid of small overlapping patches. This training tactic does not weaken the performance of network but it leads to lessen the training time. The HR patches are cropped into the squared size of 64, 128, and 256 according to the up-scaling factors of 4, 8, and 16, respectively. To produce the LR depth patches, we down-sample the HR depth patches to the fixed size of $16 \times 16$ by using the Bicubic interpolation. The metric of Mean Absolute Difference (MAD) and Root Mean Square Error (RMSE) are introduced for quantitative evaluation.

We implement our network with PyTorch and train with an NVIDIA 2080Ti GPU. We also implement our network by using the MindSpore Lite tool\footnote{https://www.mindspore.cn/}. During training, a batch size of 8 is applied. We use ADAM with momentum of 0.9, $\beta_1=0.9$, $\beta_2 = 0.99$, $\epsilon = 10^{-8}$ for network optimization. The learning rate is initiated to $1e^{-4}$, which will be decreased by multiplying 0.1 for every 100 epochs. Under $\times 8$ Depth SR, the inference time for an image with the size of $256\times 256$ is $0.052$ second via the aforementioned GPU. 

\subsection{Performance Comparison}
%\vspace{0.2cm}

\textbf{Middlebury Dataset.} We compare with some state-of-the-art DSR methods under different up-sampling factors ($\times$ 4, $\times$ 8, and $\times$ 16), including six traditional depth SR methods (\ie, CLMF \cite{LuSMLD12}, JGF \cite{0001TT13}, TGV \cite{DBLP:conf/iccv/FerstlRRRB13}, CDLLC \cite{DBLP:conf/icmcs/XieCFS14}, PB \cite{DBLP:conf/eccv/AodhaCNB12}, and EG \cite{DBLP:journals/tip/XieFS16} ) and seven deep learning based methods (\ie, SRCNN \cite{DBLP:conf/eccv/DongLHT14}, ATGVNet \cite{DBLP:conf/eccv/RieglerRB16}, MSG \cite{HuiLT16}, DGDIE \cite{DBLP:conf/cvpr/GuZGCCZ17}, DEIN \cite{DBLP:conf/icassp/YeDL18}, CCFN \cite{WenSLLF19}, GSRPT \cite{LutioDWS19}, and CTKT \cite{Sun2021cvpr}).

\begin{figure*}[!ht]
  \centering
  \begin{minipage}[b]{\linewidth} 
  \subfloat[]{
    \begin{minipage}[b]{0.23\linewidth} 
      \centering
      \includegraphics[width=\linewidth]{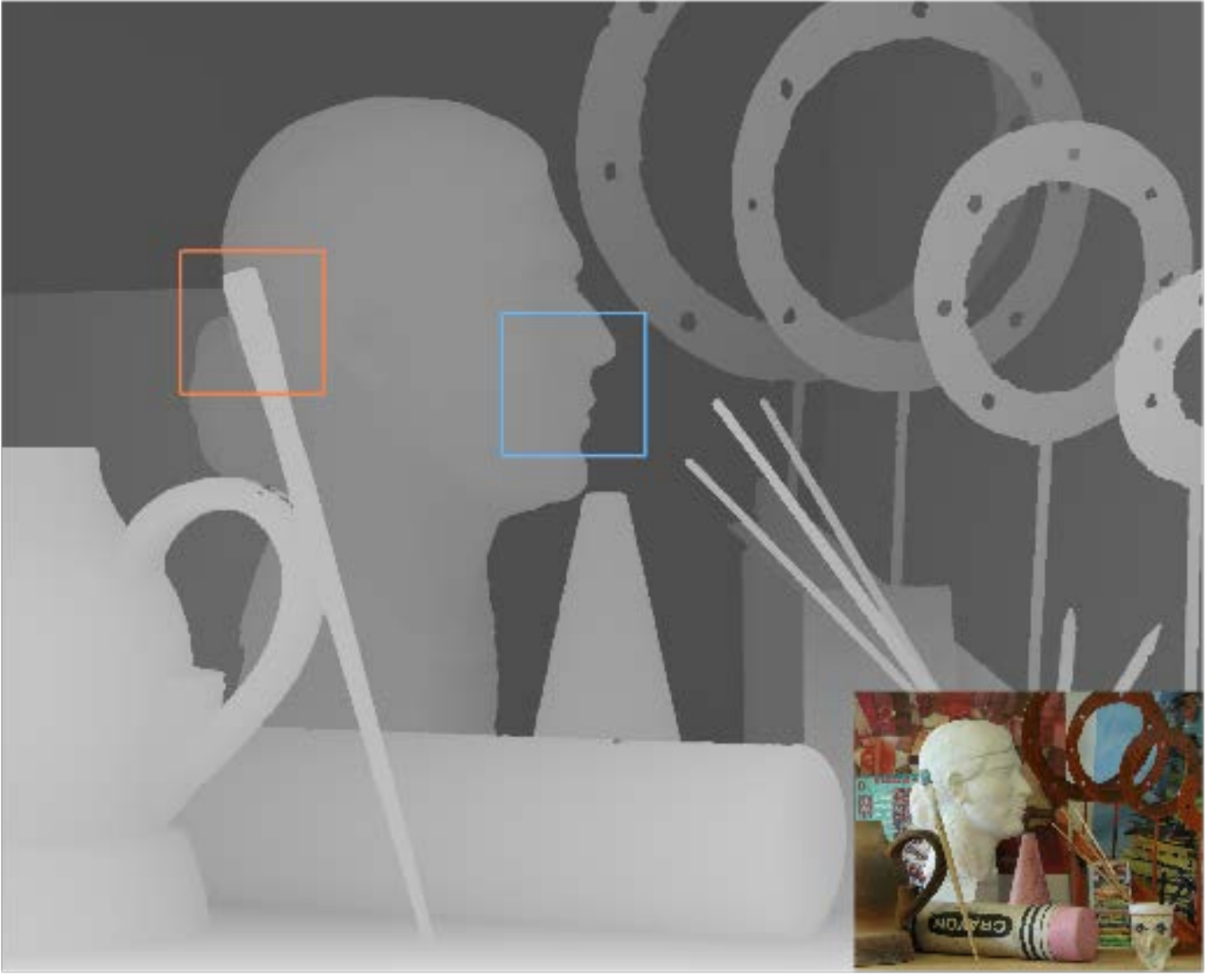}\vspace{2pt}
      \includegraphics[width=\linewidth]{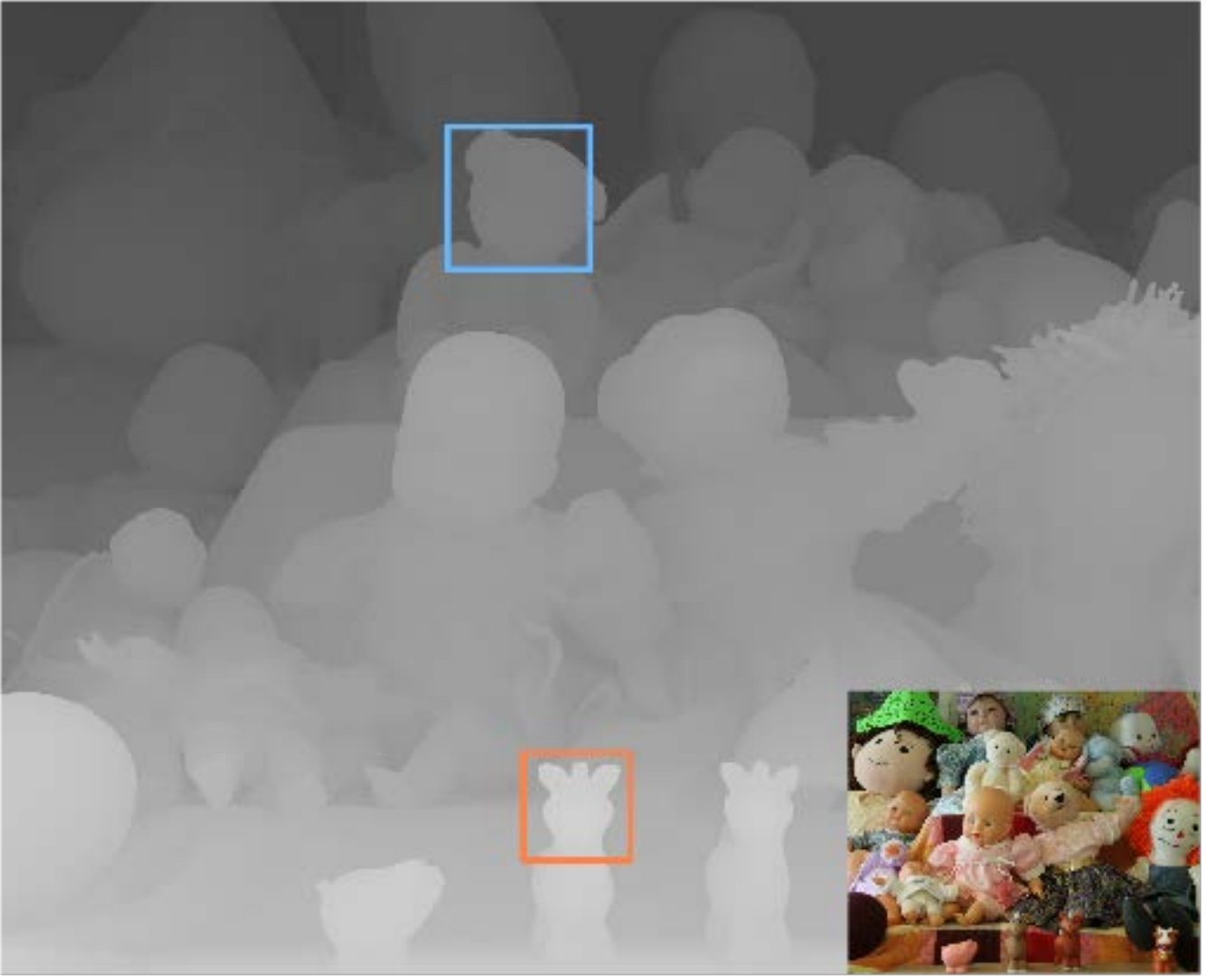}
       \end{minipage}
  }
  \hspace{-1.2mm}	
  \subfloat[]{
    \begin{minipage}[b]{0.09\linewidth} 
      \centering
      \includegraphics[width=\linewidth]{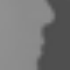}\vspace{2pt}
      \includegraphics[width=\linewidth]{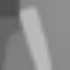}\vspace{2pt}
      \includegraphics[width=\linewidth]{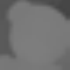}\vspace{2pt}
      \includegraphics[width=\linewidth]{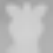}
       \end{minipage}
  }
  \hspace{-1.2mm}
    \subfloat[]{
    \begin{minipage}[b]{0.09\linewidth}
      \centering
      \includegraphics[width=\linewidth]{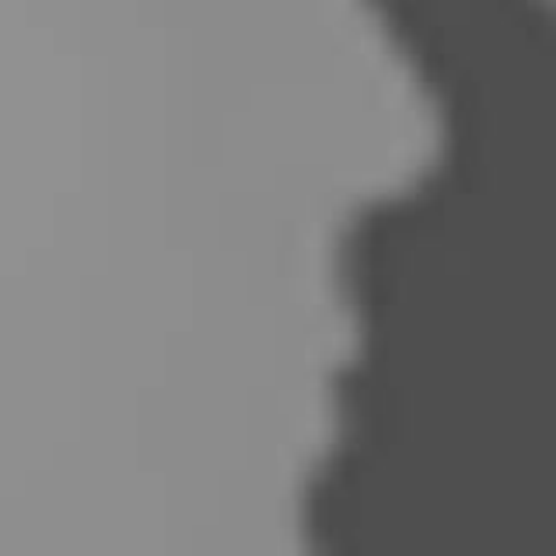}\vspace{2pt}
      \includegraphics[width=\linewidth]{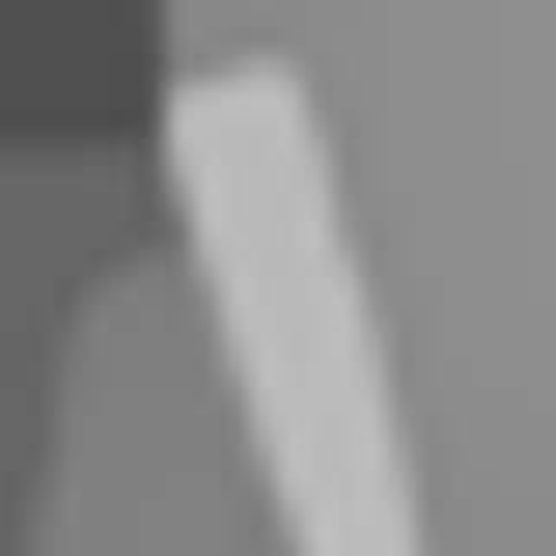}\vspace{2pt}
      \includegraphics[width=\linewidth]{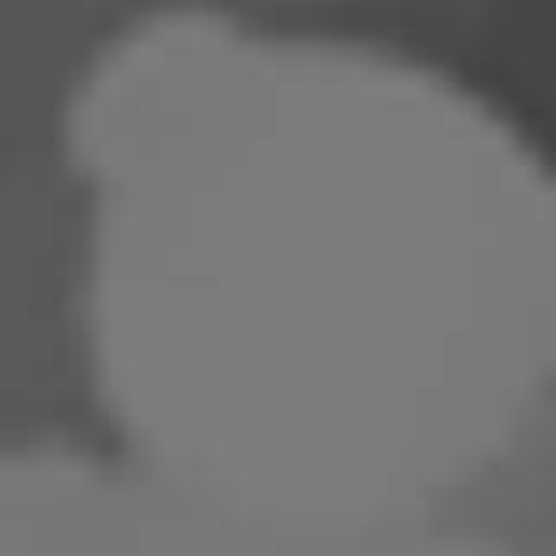}\vspace{2pt}
      \includegraphics[width=\linewidth]{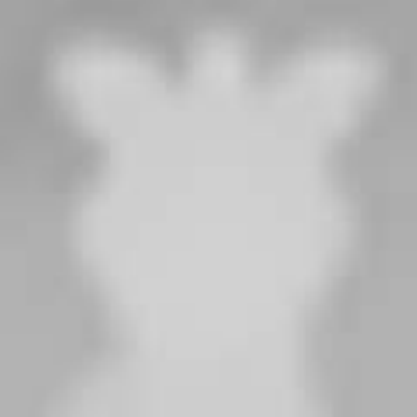}
       \end{minipage}
  }
   \hspace{-1.2mm}
   \subfloat[]{
    \begin{minipage}[b]{0.09\linewidth}
      \centering
      \includegraphics[width=\linewidth]{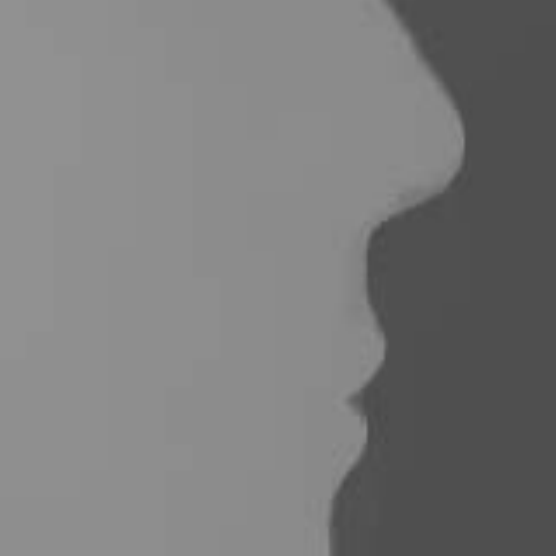}\vspace{2pt}
      \includegraphics[width=\linewidth]{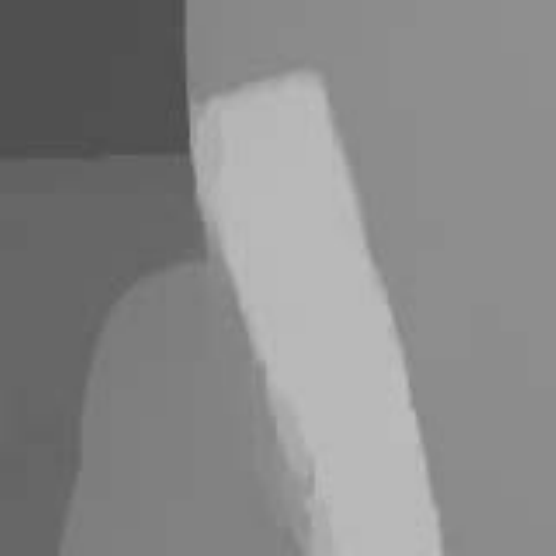}\vspace{2pt}
      \includegraphics[width=\linewidth]{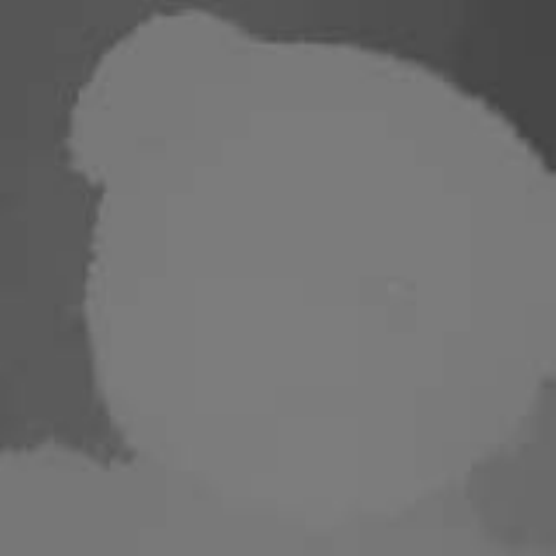}\vspace{2pt}
      \includegraphics[width=\linewidth]{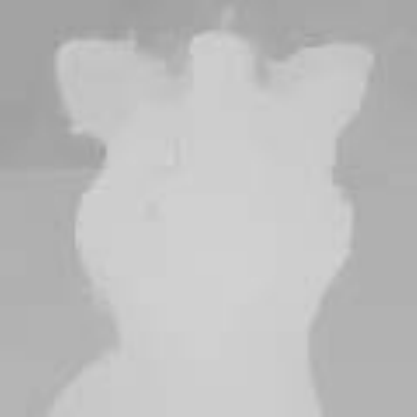}
       \end{minipage}
  }
   \hspace{-1.2mm}
  \subfloat[]{
    \begin{minipage}[b]{0.09\linewidth}
      \centering
      \includegraphics[width=\linewidth]{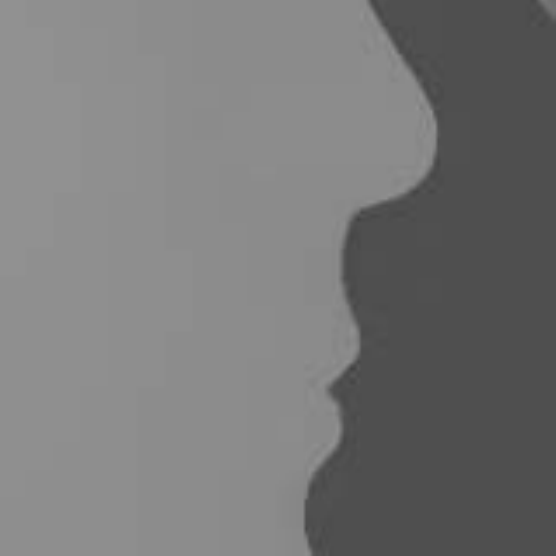}\vspace{2pt}
      \includegraphics[width=\linewidth]{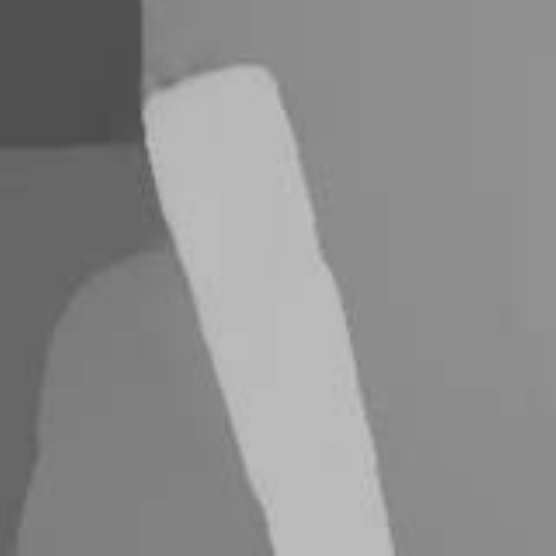}\vspace{2pt}
      \includegraphics[width=\linewidth]{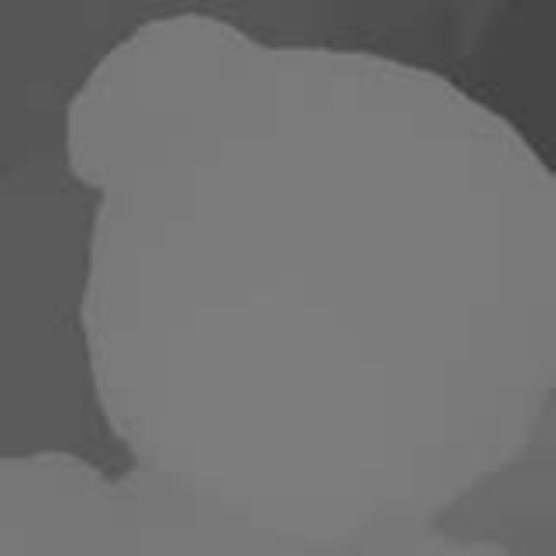}\vspace{2pt}
      \includegraphics[width=\linewidth]{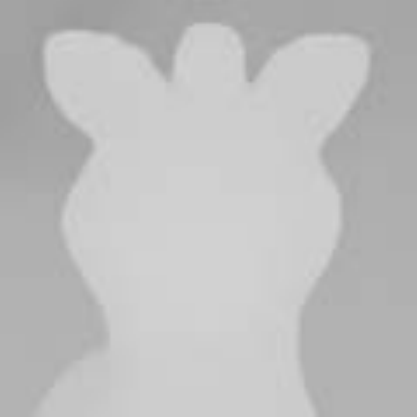}
       \end{minipage}
  }
   \hspace{-1.2mm}
   \subfloat[]{
    \begin{minipage}[b]{0.09\linewidth}
      \centering
      \includegraphics[width=\linewidth]{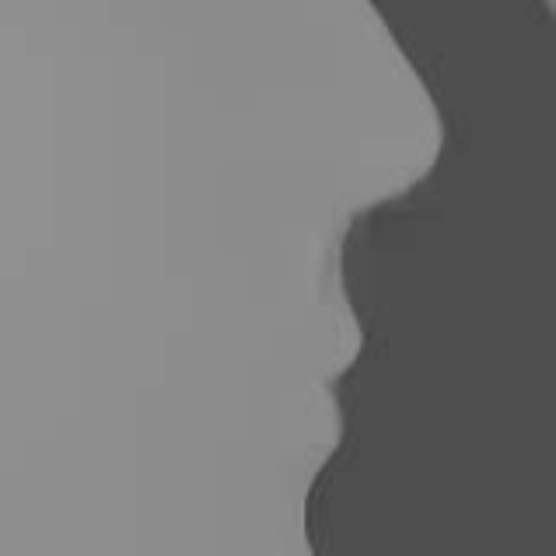}\vspace{2pt}
      \includegraphics[width=\linewidth]{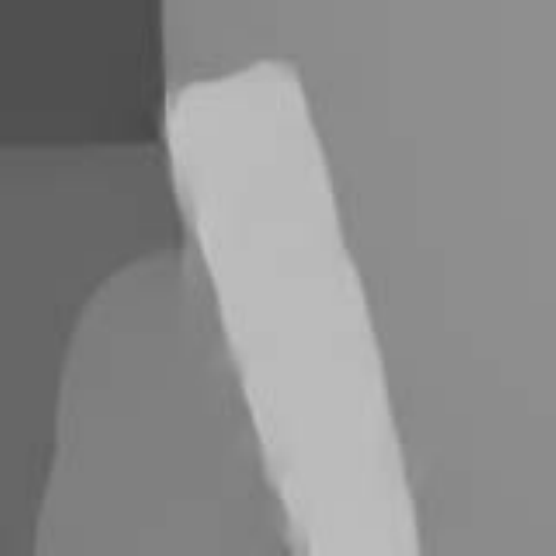}\vspace{2pt}
      \includegraphics[width=\linewidth]{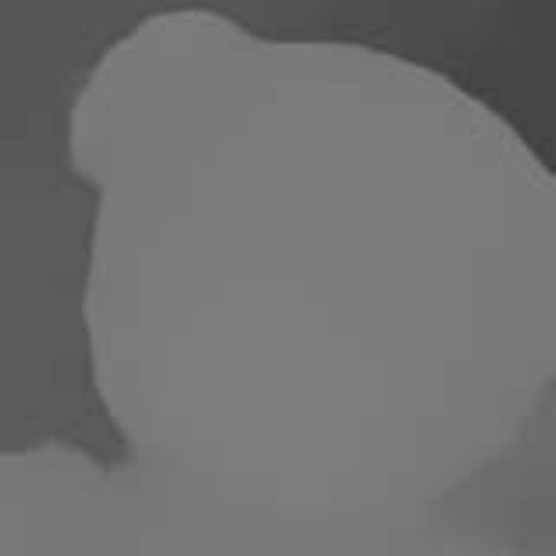}\vspace{2pt}
      \includegraphics[width=\linewidth]{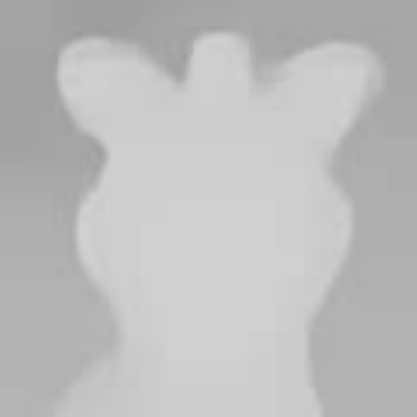}
       \end{minipage}
  }
   \hspace{-1.2mm}
  \subfloat[]{
    \begin{minipage}[b]{0.09\linewidth}
      \centering
      \includegraphics[width=\linewidth]{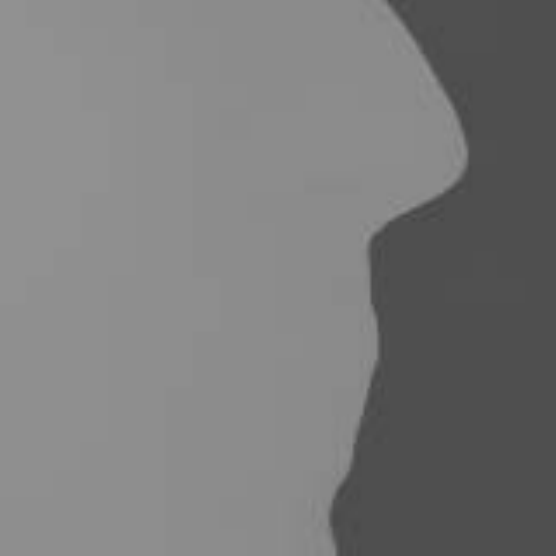}\vspace{2pt}
      \includegraphics[width=\linewidth]{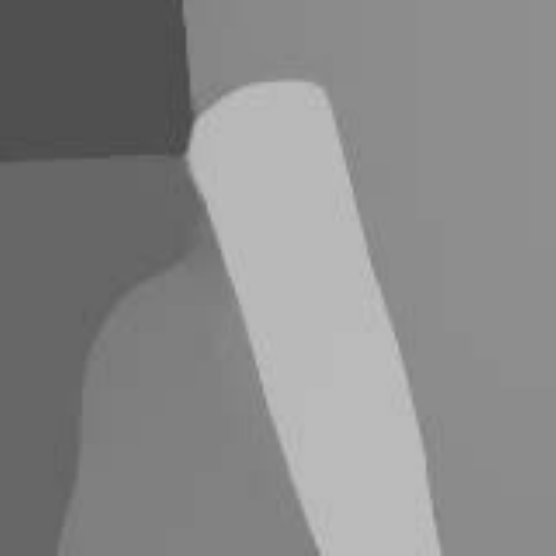}\vspace{2pt}
      \includegraphics[width=\linewidth]{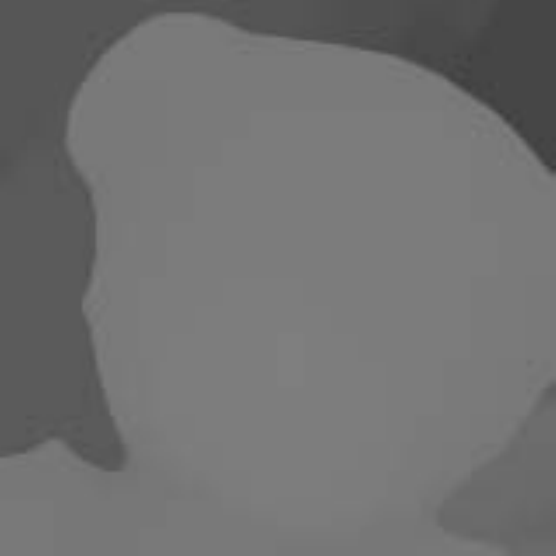}\vspace{2pt}
      \includegraphics[width=\linewidth]{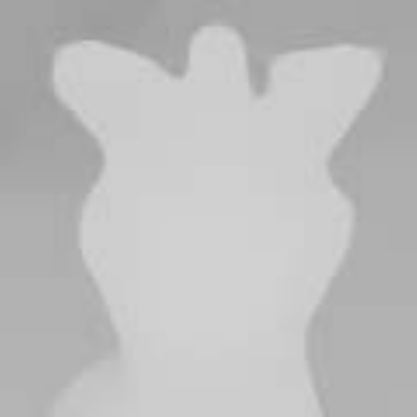}
       \end{minipage}
  }
 \hspace{-1.2mm}
  \subfloat[]{
    \begin{minipage}[b]{0.09\linewidth}
      \centering
      \includegraphics[width=\linewidth]{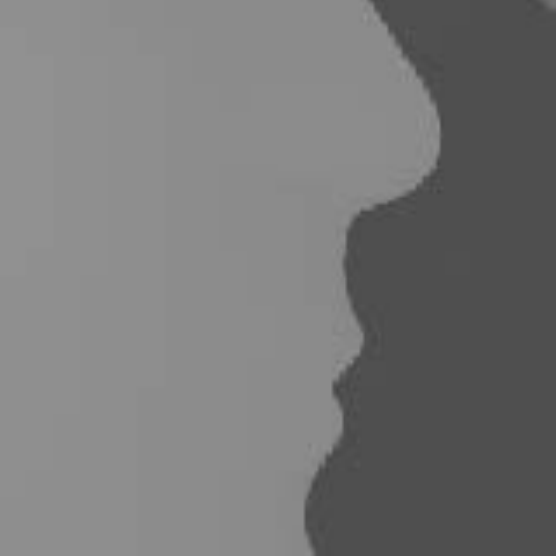}\vspace{2pt}
      \includegraphics[width=\linewidth]{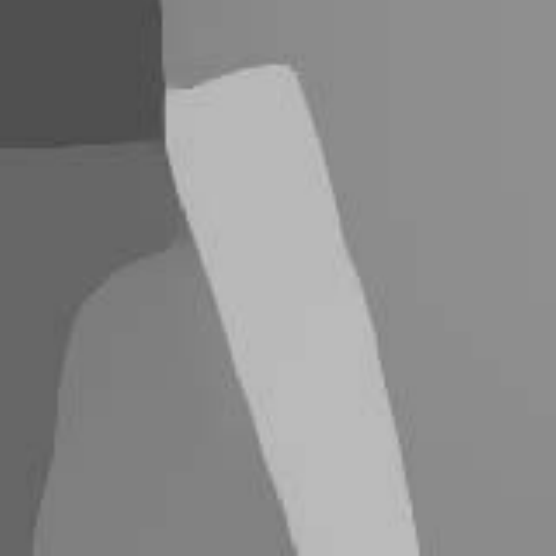}\vspace{2pt}
      \includegraphics[width=\linewidth]{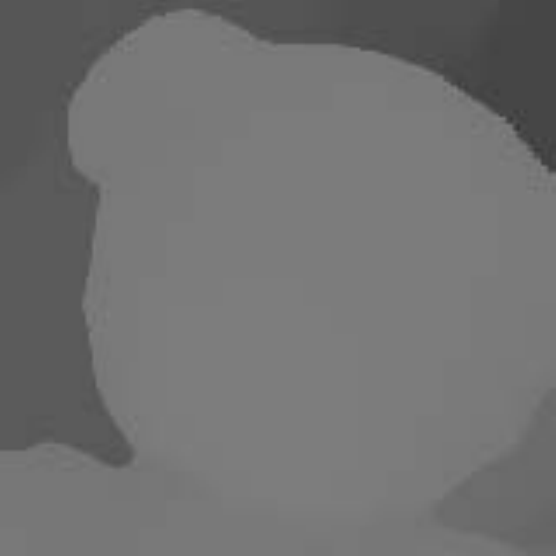}\vspace{2pt}
      \includegraphics[width=\linewidth]{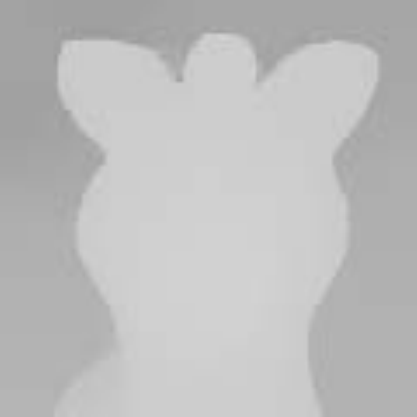}
       \end{minipage}
  }
   \hspace{-1.2mm}
    \subfloat[]{
    \begin{minipage}[b]{0.09\linewidth}
      \centering
      \includegraphics[width=\linewidth]{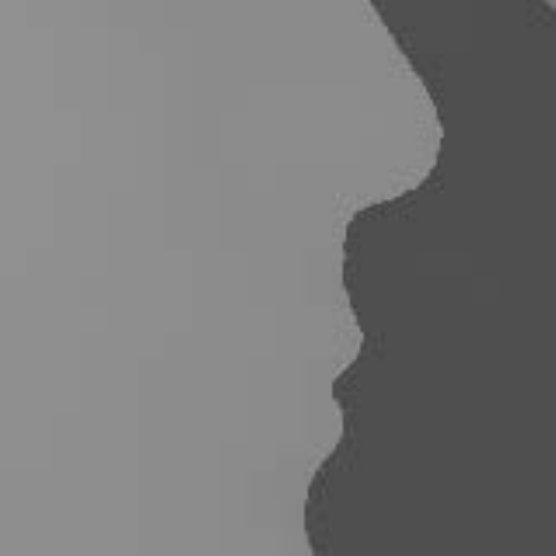}\vspace{2pt}
      \includegraphics[width=\linewidth]{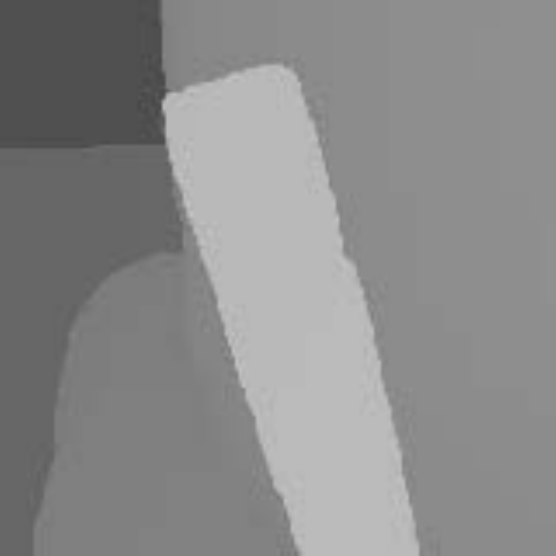}\vspace{2pt}
      \includegraphics[width=\linewidth]{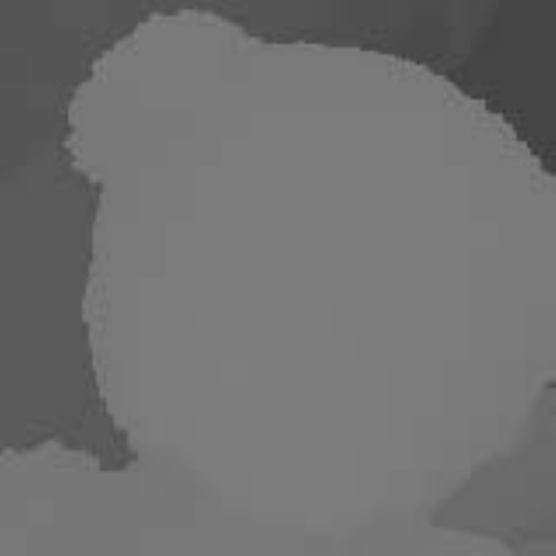}\vspace{2pt}
      \includegraphics[width=\linewidth]{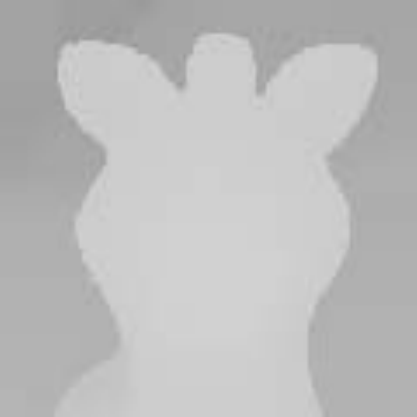}
       \end{minipage}
  }
  \end{minipage}
  \vfill
  \caption{Visual comparisons of $\times 8$ up-sampling results on two examples (\ie, Art in the first row and Dolls in the second row). (a) Ground truth depth maps and color images; (b) LR depth patches; (c)-(h) The super-resolved depth maps generated by Bicubic, TGV \cite{DBLP:conf/iccv/FerstlRRRB13}, MSG \cite{HuiLT16}, DGDIE \cite{DBLP:conf/cvpr/GuZGCCZ17}, CTKT \cite{Sun2021cvpr}, and BridgeNet, respectively. (i) Ground truth. Depth patches are enlarged for clear visualization.}
  \label{fig:fig_7}
\vspace{-0.225cm}
\end{figure*}

\begin{table*}[!htbp]
\caption{Quantitative depth SR results (in MAD) on Middlebury 2005 dataset. The best performance is displayed in bold, and the second best performance is marked in underline.}
\label{tab:tab1}
\renewcommand{\arraystretch}{0.85}
		\setlength{\tabcolsep}{1.4mm}{
\begin{tabular}{c|ccc|ccc|ccc|ccc|ccc|ccc}
\hline
\multirow{2}{*}{} & \multicolumn{3}{c|}{Art} & \multicolumn{3}{c|}{Books} & \multicolumn{3}{c|}{Dolls} & \multicolumn{3}{c|}{Laundry} & \multicolumn{3}{c|}{Mobius} & \multicolumn{3}{c}{Reindeer} \\
\cline{2-19}
 &$\times 4$  &$\times 8$  &  $\times 16$& $\times 4$ & $\times 8$ & $\times 16$ & $\times 4$ & $\times 8$ & $\times 16$ & $\times 4$ & $\times 8$ & $\times 16$ & $\times 4$ & $\times 8$ & $\times 16$ & $\times 4$ & $\times 8$ & $\times 16$ \\
 \hline
 CLMF \cite{LuSMLD12}& 0.76 & 1.44 & 2.87 & 0.28 & 0.51 & 1.02 & 0.34 & 0.60 & 1.01 & 0.50 & 0.80 & 1.67 & 0.29 & 0.51 & 0.97 & 0.51 & 0.84 & 1.55 \\
 JGF \cite{0001TT13}& 0.47 & 0.78 & 1.54 & 0.24 & 0.43 & 0.81 & 0.33 & 0.59 & 1.06 & 0.36 & 0.64 & 1.20 & 0.25 & 0.46 & 0.80 & 0.38 & 0.64 & 1.09\\
% EDGE& 0.65 & 1.03 & 2.11 & 0.30 & 0.56 & 1.03 & 0.31 & 0.56 & 1.05 & 0.32 & 0.54 & 1.14 & 0.29 & 0.51 & 1.10 & 0.37 & 0.63 & 1.28\\
 TGV \cite{DBLP:conf/iccv/FerstlRRRB13}& 0.65 & 1.17 & 2.30 & 0.27 & 0.42 & 0.82 & 0.33 & 0.70 & 2.20 & 0.55 & 1.22 & 3.37 & 0.29 & 0.49 & 0.90 & 0.49 & 1.03 & 3.05\\
% KSVD& 0.64 & 0.81 & 1.47 & 0.23 & 0.52 & 0.76 & 0.34 & 0.56 & 0.82 & 0.35 & 0.52 & 1.08 & 0.28 & 0.48 & 0.81 & 0.47 & 0.57 & 0.99\\
 CDLLC \cite{DBLP:conf/icmcs/XieCFS14}& 0.53 & 0.76 & 1.41 & 0.19 & 0.46 & 0.75 & 0.31 & 0.53 & 0.79 & 0.30 & 0.48 & 0.96 & 0.27 & 0.46 & 0.79 & 0.43 & 0.55 & 0.98\\
 PB \cite{DBLP:conf/eccv/AodhaCNB12}& 0.79 & 0.93 & 1.98 & 0.16 & 0.43 & 0.79 & 0.53 & 0.83 & 0.99 & 1.13 & 1.89 & 2.87 & 0.17 & 0.47 & 0.82 & 0.56 & 0.97 & 1.89\\
 EG \cite{DBLP:journals/tip/XieFS16}& 0.48 & 0.71 & \underline{1.35} & 0.15 & 0.36 & 0.70 & 0.27 & 0.49 & 0.74 & 0.28 & 0.45 & 0.92 & 0.23 & 0.42 & 0.75 & 0.36 & 0.51 & 0.95\\\hline
 SRCNN \cite{DBLP:conf/eccv/DongLHT14}& 0.63 & 1.21 & 2.34 & 0.25 & 0.52 & 0.97 & 0.29 & 0.58 & 1.03 & 0.40 & 0.87 & 1.74 & 0.25 & 0.43 & 0.87 & 0.35 & 0.75 & 1.47 \\
% DSP& 0.73 & 1.56 & 3.03 & 0.28 & 0.61 & 1.31 & 0.32 & 0.65 & 1.45 & 0.45 & 0.98 & 2.01 & 0.31 &0.59  & 1.26 & 0.42 & 0.84 &1.73 \\
 ATGVNet \cite{DBLP:conf/eccv/RieglerRB16}& 0.65 & 0.81 & 1.42 & 0.43 & 0.51 & 0.79 &0.41  & 0.52 & \textbf{0.56} & 0.37 & 0.89 & 0.94 & 0.38 &  0.45& 0.80 & 0.41 & 0.58 &1.01 \\
 MSG \cite{HuiLT16}& 0.46 & 0.76 & 1.53 & 0.15 & 0.41 & 0.76 & 0.25 & 0.51 & 0.87 & 0.30 & 0.46 & 1.12 & 0.21 & 0.43 & 0.76 & 0.31 & 0.52 & 0.99\\
 DGDIE \cite{DBLP:conf/cvpr/GuZGCCZ17}& 0.48 & 1.20 &2.44  & 0.30 & 0.58 & 1.02 & 0.34 & 0.63 & 0.93 & 0.35 & 0.86 & 1.56 & 0.28 & 0.58 & 0.98 & 0.35 & 0.73 &1.29 \\
 DEIN \cite{DBLP:conf/icassp/YeDL18}& 0.40 & 0.64 & \textbf{1.34} & 0.22 & 0.37 & 0.78 & 0.22 & 0.38 & 0.73 & 0.23 & \underline{0.36} & 0.81 & 0.20 & 0.35 & 0.73 & 0.26 & 0.40 & 0.80\\
 CCFN \cite{WenSLLF19}& 0.43 & 0.72 & 1.50 & 0.17 & 0.36 & 0.69 & 0.25 & 0.46 & 0.75 & 0.24 & 0.41 & \textbf{0.71} & 0.23 & 0.39 & 0.73 & 0.29 & 0.46 & 0.95\\
 GSRPT \cite{LutioDWS19}& 0.48 & 0.74 & 1.48 & 0.21 & 0.38 & 0.76 & 0.28 & 0.48 & 0.79 & 0.33 & 0.56 & 1.24 & 0.24 & 0.49 &  0.80& 0.31 & 0.61 &1.07 \\
CTKT \cite{Sun2021cvpr}& \textbf{0.25} & \textbf{0.53} & 1.44 & \textbf{0.11} & \underline{0.26} & \underline{0.67} & \textbf{0.16} & \underline{0.36} & 0.65 & \textbf{0.16} & \underline{0.36} & \underline{0.76} & \textbf{0.13} & \underline{0.27} & \underline{0.69} & \textbf{0.17} & \underline{0.35} & \underline{0.77}\\
BridgeNet (Ours)& \underline{0.30} & \underline{0.58} & 1.49 & \underline{0.14} & \textbf{0.24} & \textbf{0.51} &\underline{0.19}  & \textbf{0.34} & \underline{0.64} & \underline{0.17} & \textbf{0.34} & \textbf{0.71} & \underline{0.15} & \textbf{0.26} & \textbf{0.54} & \underline{0.19} & \textbf{0.31} & \textbf{0.70}\\
 \hline
\end{tabular}}
\vspace{-0.3cm}
\end{table*}

Figure \ref{fig:fig_7} demonstrates the visual comparisons of different methods under the factor of $\times 8$. As visible, our method can recover more fine-grained depth details, \eg, less artifacts around the stick in the Art image, more accurate shape of the toy head in the Dolls image. The comparison methods may generate some artifacts, blurred boundaries, or shape changes. Those phenomena are related to the severe damages on fine structure and tiny objects during down-sampling degradation, which brings more difficulties on these regions. In contrast, our method has advantages in accurately recovering the depth boundaries of these tiny objects.
The quantitative comparisons are reported in Table \ref{tab:tab1}, we can see that our network obtains the competitive result against other comparison methods, even under challenging scaling factors of $\times$8 and $\times$16. Taking the $\times$16 SR as an example, for the Books image, compared with the \textbf{\emph{second best}} algorithm, our method refreshes the MAD from 0.67 to 0.51, with the percentage gain of 23.9\%.

\begin{figure*}[!ht]
\centering
\centerline{\includegraphics[width=1\textwidth]{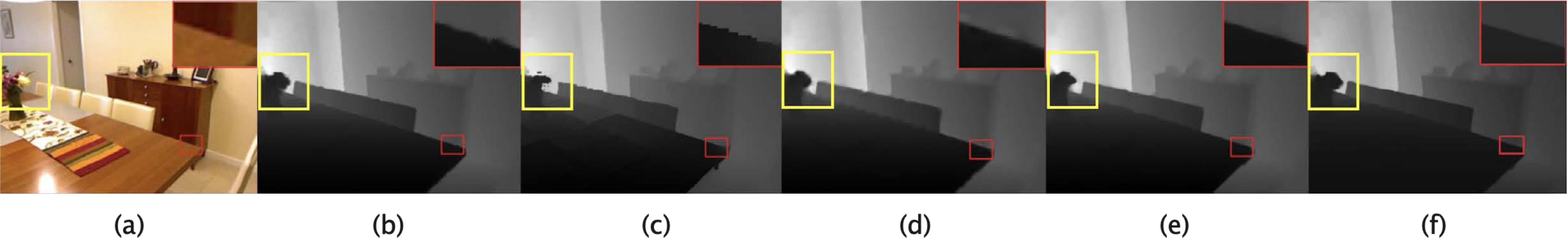}}
\caption{Visual comparisons of different method under $\times 8$ up-sampling on the NYU v2 dataset. (a) Color image. (b) Ground truth. (c) SDF \cite{DBLP:journals/pami/HamCP18}. (d) DJF \cite{DBLP:conf/eccv/LiHA016}. (e) SVLRM \cite{DBLP:conf/cvpr/PanDRLT019}. (f) BridgeNet.
}
\label{fig:fig_8}
 \vspace{-0.225cm}
\end{figure*}

\textbf{NYU v2 Dataset.} We also evaluate our method on the NYU v2 dataset and compare it with other SOTA methods, including Bicubic, TGV \cite{DBLP:conf/iccv/FerstlRRRB13}, EDGE \cite{ParkKTBK11}, DJF \cite{DBLP:conf/eccv/LiHA016}, SDF \cite{DBLP:journals/pami/HamCP18}, DGDIE \cite{DBLP:conf/cvpr/GuZGCCZ17}, GbFT \cite{DBLP:conf/iccv/AlbaharH19}, PAC \cite{DBLP:conf/cvpr/SuJSGLK19}, SVLRM \cite{DBLP:conf/cvpr/PanDRLT019}, DKN \cite{DBLP:journals/corr/abs-1903-11286}, and CTKT \cite{Sun2021cvpr}.
Figure \ref{fig:fig_8} demonstrates the visual results of our method under the $\times 8$ up-sampling. Whether in the red or yellow rectangular area, our model can accurately reconstruct the depth information and edges of the small objects. As reported in Table \ref{tab:tab2}, our method achieves the best performance under the $\times 8$ and $\times 16$ up-sampling cases. Compared with the \emph{\textbf{second best}} algorithm, the RMSE of our method reaches 2.46 under the scaling factor of $\times 8$, with an improvement of 9.9\%.

\begin{table*}[!ht]
\caption{Quantitative depth SR results (in RMSE) on NYU v2 dataset. The best performance is displayed in bold, and the second best performance is marked in underline.}
\centering
\label{tab:tab2}
\setlength{\tabcolsep}{0.5mm}{
\begin{tabular}{c|cccccccccccc}
\hline
 & Bicubic & TGV \cite{DBLP:conf/iccv/FerstlRRRB13} & EDGE \cite{ParkKTBK11} & DJF \cite{DBLP:conf/eccv/LiHA016}&SDF \cite{DBLP:journals/pami/HamCP18} & DGDIE \cite{DBLP:conf/cvpr/GuZGCCZ17}& GbFT \cite{DBLP:conf/iccv/AlbaharH19} & PAC \cite{DBLP:conf/cvpr/SuJSGLK19}& SVLRM \cite{DBLP:conf/cvpr/PanDRLT019}& DKN \cite{DBLP:journals/corr/abs-1903-11286}& CTKT \cite{Sun2021cvpr} & BridgeNet (Ours)\\
\hline
$\times 4$ & 8.16 & 6.98 & 5.21 & 3.54 & 3.04&1.56&3.35&2.39&1.74&1.62&\textbf{1.49} & \underline{1.54}\\

$\times 8$ & 14.22 & 11.23 & 9.56 & 6.20 &5.67 &2.99&5.73&4.59&5.59&3.26&\underline{2.73} & \textbf{2.63} \\

$\times 16$ & 22.32 & 28.13 & 18.10 & 10.21 &9.97 &5.24&9.01&8.09&7.23&6.51&\underline{5.11} & \textbf{4.98}\\
\hline
\end{tabular}}
\vspace{-0.3cm}
\end{table*}

\subsection{Ablation Study}
%\vspace{0.2cm}
%of our interaction between DSR and MDE, \ie, High-Frequency Attention Bridge (HABdg) and Content-Guidance Bridge (CGBdg) by ablation study.

\begin{table}[!ht]
\caption{Ablation studies (in MAD) of our BridgeNet on the Middlebury 2005 dataset ($\times 8$ case).}
\centering
\label{tab:tab3}
\setlength{\tabcolsep}{2mm}{
\begin{tabular}{c|cccc|c}
\hline
& DSRNet & MDENet & HABdg & CGBdg & Middlebury\\
\hline
1 &\checkmark & & &  & 0.366\\
2& & \checkmark &  & & 0.472\\
 \hline
3&\checkmark & \checkmark  & & & 0.363\\
4&\checkmark & \checkmark  &\checkmark & & 0.355\\
5&\checkmark & \checkmark  &  & \checkmark & 0.361\\
\hline
6&\checkmark & \checkmark & \checkmark & \checkmark & 0.343\\
\hline
\end{tabular}}
\end{table}

\begin{figure}[!ht]
\vspace{-0.4cm}
  \centering
  \begin{minipage}[b]{\linewidth} 
  \subfloat[]{
    \begin{minipage}[b]{0.3\linewidth} 
      \centering
      \includegraphics[width=\linewidth]{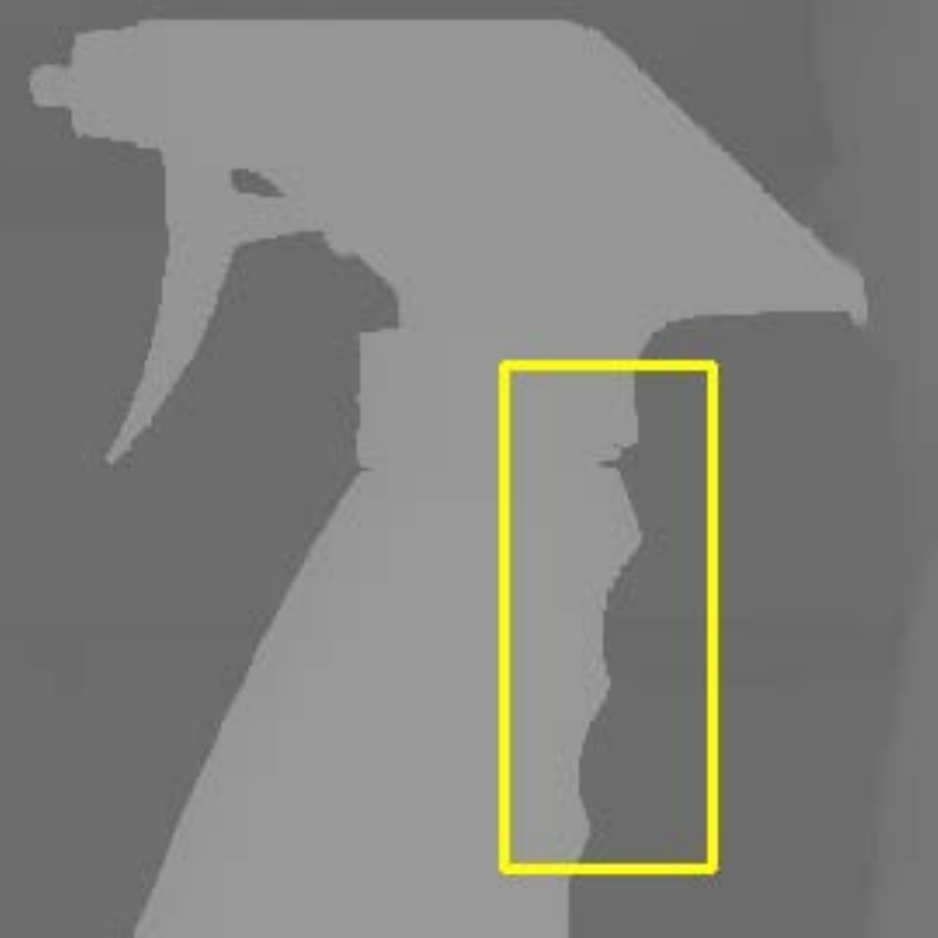}\vspace{2pt}
      \includegraphics[width=\linewidth]{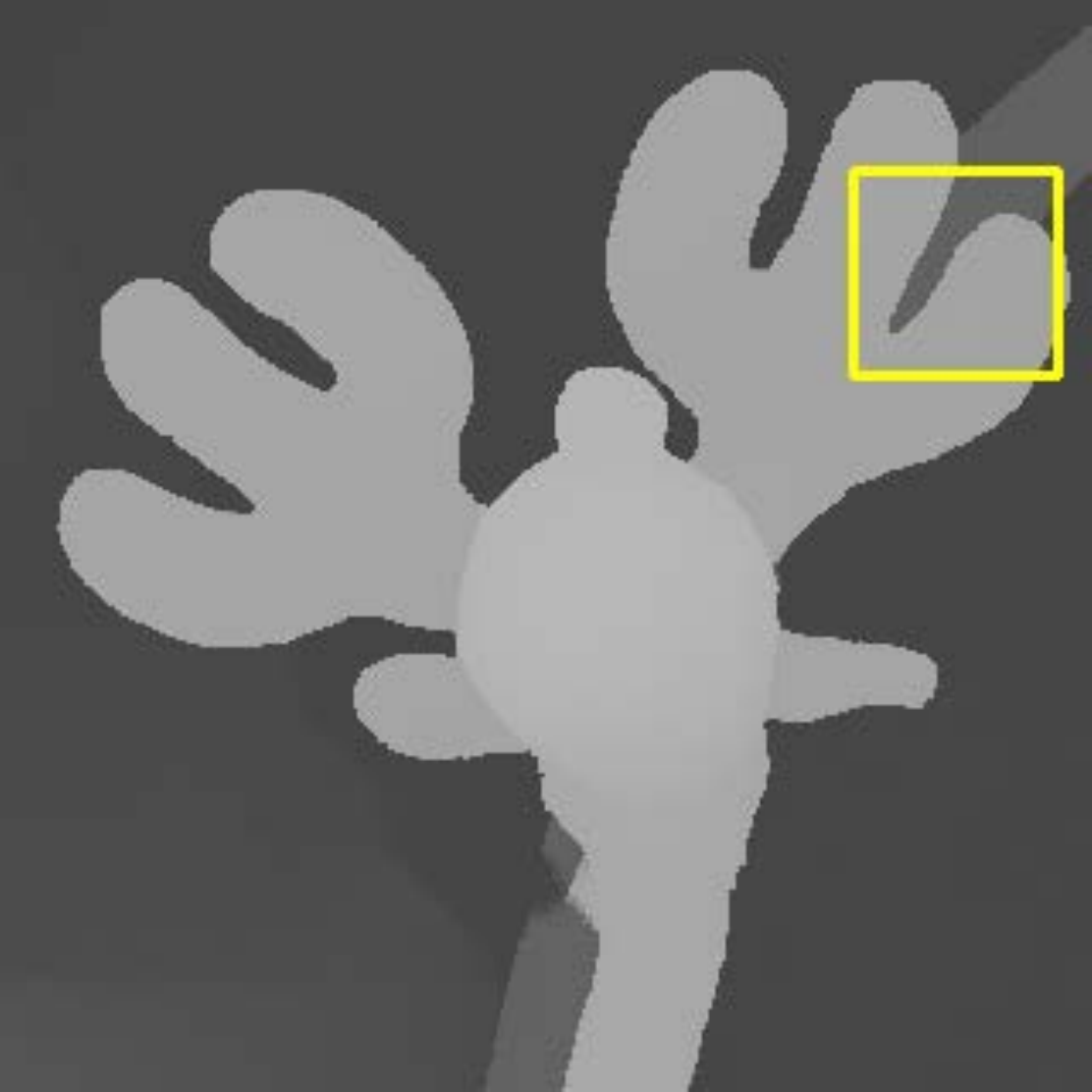}
     \end{minipage}
  }
  \hfill
    \subfloat[]{
    \begin{minipage}[b]{0.3\linewidth}
      \centering
      \includegraphics[width=\linewidth]{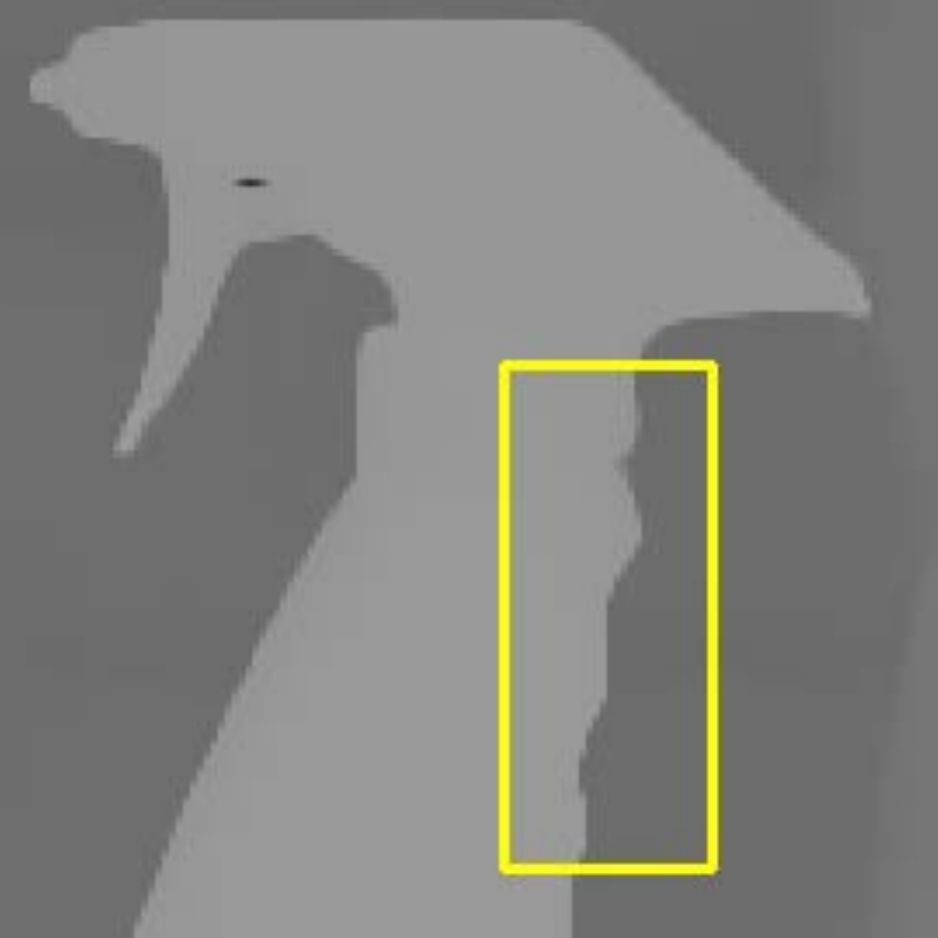}\vspace{2pt}
      \includegraphics[width=\linewidth]{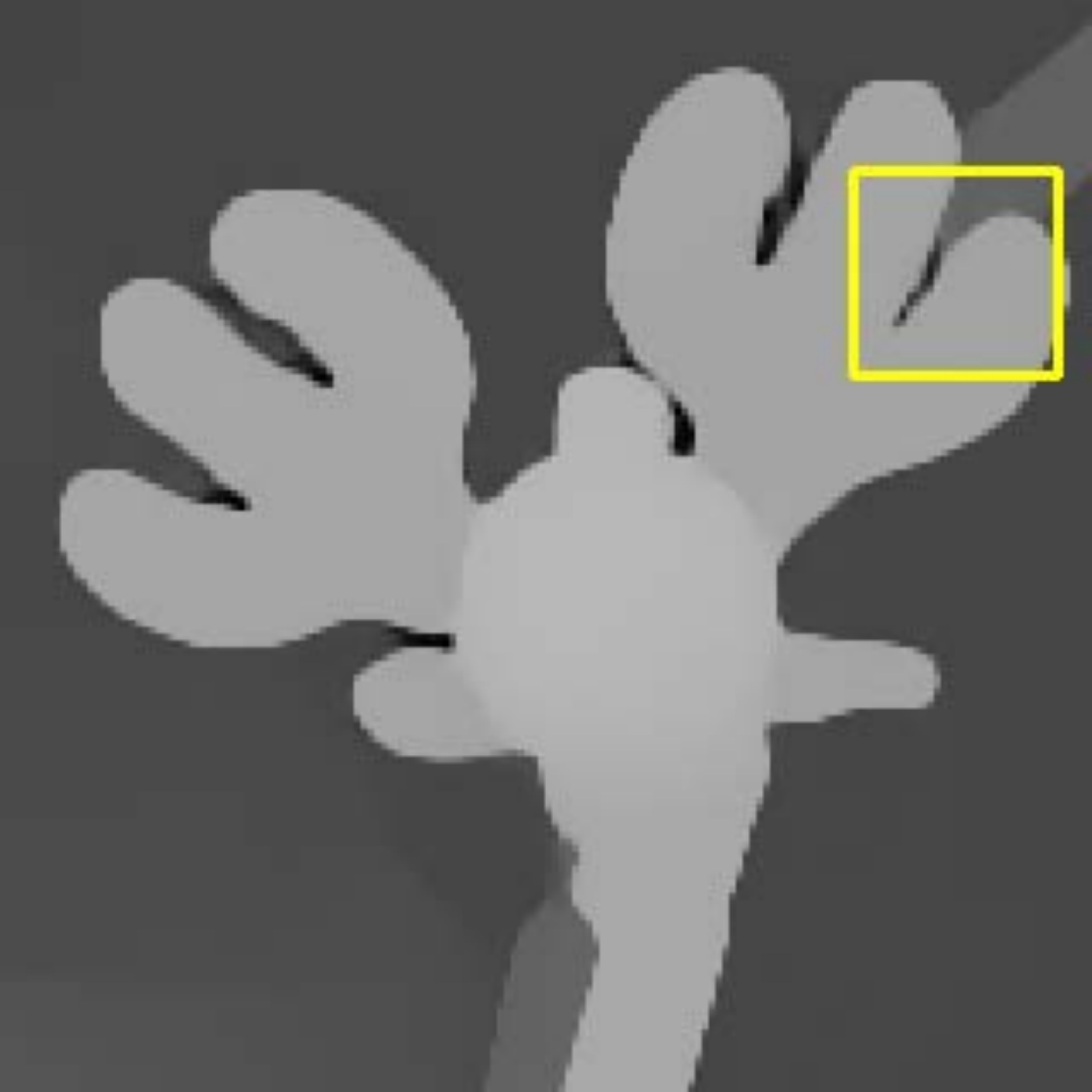}
       \end{minipage}
  }
   \hfill
   \subfloat[]{
    \begin{minipage}[b]{0.3\linewidth}
      \centering
      \includegraphics[width=\linewidth]{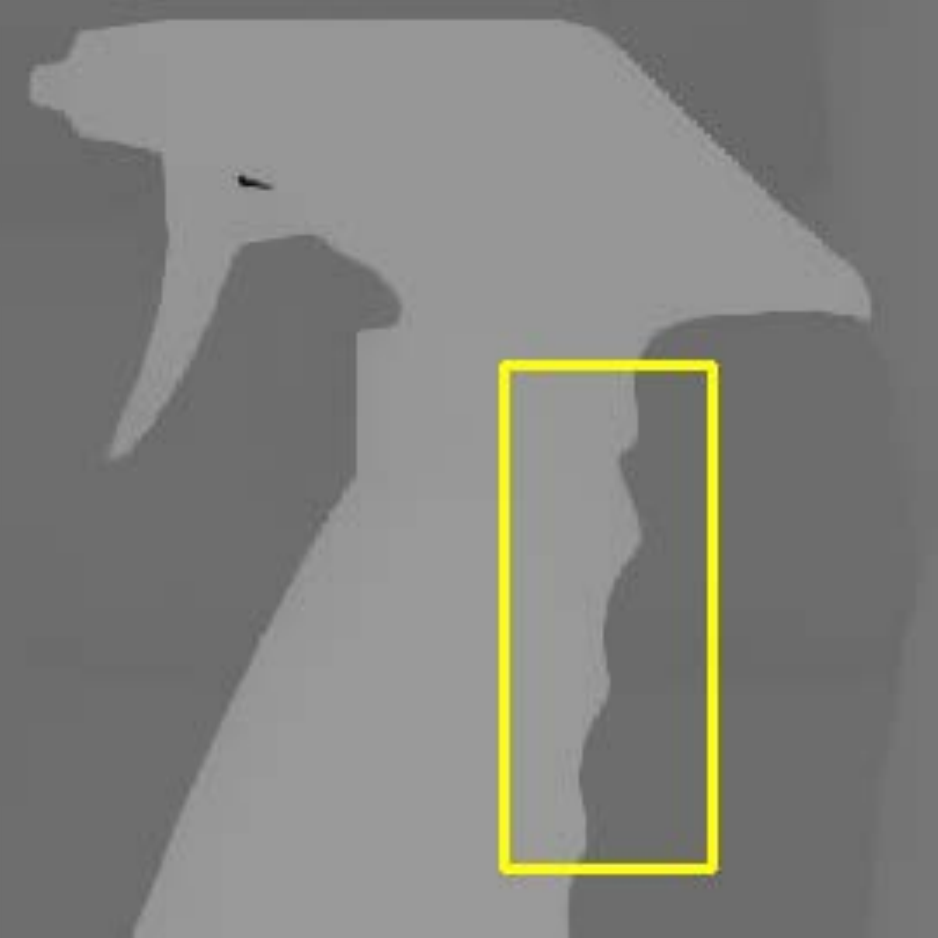}\vspace{2pt}
      \includegraphics[width=\linewidth]{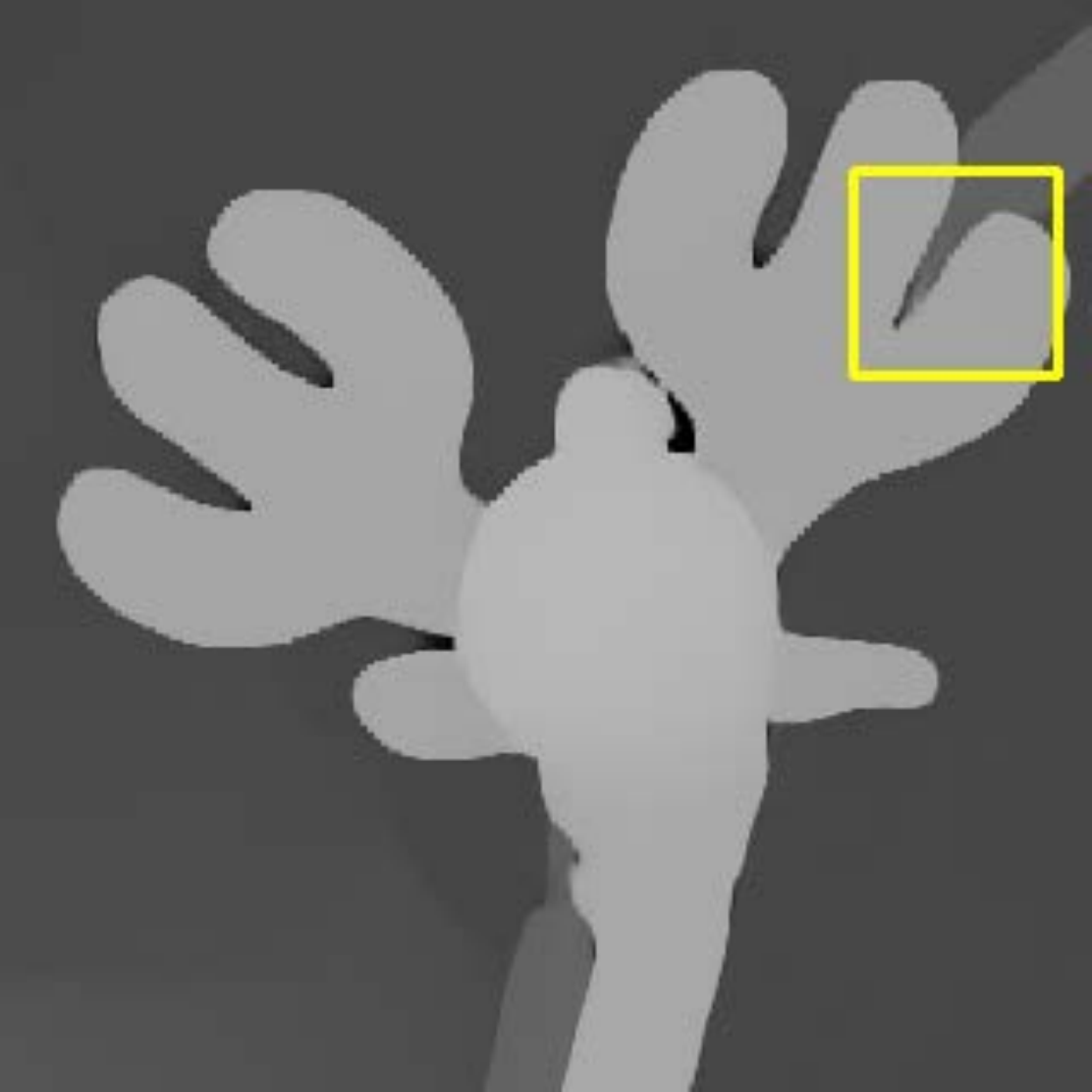}
       \end{minipage}
  }
  \end{minipage}
  \vfill
  \caption{Visual comparisons of different components of our network ($\times 8$ case). (a) Ground truth. (b) DSRNet. (c) BridgeNet.}
  \label{fig:fig_9}
\vspace{-0.4cm}
\end{figure}

In this section, we conduct comprehensive ablation studies to verify the designs in our BridgeNet. We report the $\times 8$ depth SR results on Middlebury dataset under different experimental settings in Table \ref{tab:tab3}. The $1^{st}$ and $2^{nd}$ rows are the results of DSRNet and MDENet baseline trained separately without any interaction under the same settings. The performance of MDENet is inferior to the DSRNet, which coincides with the previous analysis, that is, MDE task is more difficult than DSR. 
For the multi-task interaction, we first combine the two task through the simple loss function constraint, and the result is shown in the third row of Table \ref{tab:tab3}. Compared with the DSRNet alone, the MAD of depth SR is improved to 0.363. Then, we gradually integrate HABdg and CGBdg into the framework to discuss their roles. When only using HABdg or CGBdg in the joint learning framework, results obtained are better than only using loss constraints. Moreover, when the two bridges work together (\ie, full model), the performance reaches the best. We also provide some visual comparisons in Figure \ref{fig:fig_9}. From it, we can see that our model has clearer boundaries and more accurate depth values compared with DSRNet alone, as shown in the boxes of the figure.

To further demonstrate the effectiveness of HABdg, we replace it by feeding features of MDENet into DSRNet without any processing, as presented in Table \ref{tab:tab4}. Compared with simply concatenating features in channel dimension and sending into DSRNet, our proposed HABdg can provide more effective high-frequency guidance for boosting the performance.

\begin{table}[!ht]
\caption{Ablation studies (in MAD) of our HABdg on the Middlebury 2005 dataset ($\times 8$ case). `w/o HABdg' refers to replacing the HABdg by directly propagating features from MDENet to DSRNet.}
\label{tab:tab4}
\setlength{\tabcolsep}{0.7mm}{

\begin{tabular}{c|cccccc|c}
\hline
 & Art & Books & Dolls & Laundry & Mobius & Reindeer  & Avg.\\
 \hline
w/ HABdg & 0.58 & 0.24 & 0.34 & 0.34 & 0.26 & 0.31 &0.343\\
w/o HABdg & 0.65 & 0.25 & 0.37 & 0.38 & 0.28 & 0.33& 0.376\\
\hline
\end{tabular}}
\vspace{-0.4cm}
\end{table}

\section{Conclusion}
%\vspace{0.2cm}

In this paper, we explore a joint learning framework that combines the depth map super-resolution and monocular depth estimation to boost the depth SR performance without adding any other supervision labels. The center of this paper is how to design the interaction between the two subnetworks (\ie, DSRNet and MDENet), thus we propose two bridges. On one hand, we let the MDENet provide high-frequency guidance information for the DSRNet through the HABdg in the feature encoder. On the other hand, we use the DSR branch to provide content guidance for depth estimation branch via the CGBdg in the feature decoder. Comprehensive experiments show that our method achieves competitive performance, especially for the cases of large scaling factors.
Moreover, our network architecture is highly portable and can provide a paradigm for associating the DSR and MDE tasks.

\section{Acknowledgments}
%\vspace{0.2em}

This work was supported by the Beijing Nova Program under Grant Z201100006820016, in part by the National Key Research and Development of China under Grant 2018AAA0102100, in part by the National Natural Science Foundation of China under Grant 62002014, Grant U1936212, in part by Elite Scientist Sponsorship Program by the China Association for Science and Technology under Grant 2020QNRC001, in part by General Research Fund-Research Grants Council (GRF-RGC) under Grant 9042816 (CityU 11209819), Grant 9042958 (CityU 11203820), in part by Hong Kong Scholars Program, in part by CAAI-Huawei MindSpore Open Fund, and in part by China Postdoctoral Science Foundation under Grant 2020T130050, Grant 2019M660438.

%\begin{acks}
%To Robert, for the bagels and explaining CMYK and color spaces.
%\end{acks}

%\clearpage

\bibliographystyle{ACM-Reference-Format}
\bibliography{sample-base}

%\appendix

% \end{sloppypar}
\end{document}